%% file: main.tex
\documentclass[
reprint,
superscriptaddress,
nofootinbib,
 amsmath,amssymb,
 aps,
]{revtex4-2}

\input{preamble}

\begin{document}

\title{Implicit Machine Learning Force Fields\\Accelerate Molecular Dynamics Simulations}

\author{Johannes Maeß}
\affiliation{BIFOLD, Berlin, Germany.}
\affiliation{Machine Learning Group, TU Berlin, Berlin, Germany.}
\author{Leon Werner}
\affiliation{BIFOLD, Berlin, Germany.}
\affiliation{Machine Learning Group, TU Berlin, Berlin, Germany.}
\author{J. Thorben Frank}
\affiliation{BIFOLD, Berlin, Germany.}
\affiliation{Machine Learning Group, TU Berlin, Berlin, Germany.}
\author{Winfried Ripken}
\affiliation{BIFOLD, Berlin, Germany.}
\affiliation{Machine Learning Group, TU Berlin, Berlin, Germany.}
\author{Martin Michajlow}
\affiliation{BIFOLD, Berlin, Germany.}
\affiliation{Machine Learning Group, TU Berlin, Berlin, Germany.}
\author{Joshua Futterer}
\affiliation{BIFOLD, Berlin, Germany.}
\affiliation{Machine Learning Group, TU Berlin, Berlin, Germany.}
\author{Klaus-Robert Müller}
\thanks{Corresponding authors.}
\affiliation{BIFOLD, Berlin, Germany.}
\affiliation{Machine Learning Group, TU Berlin, Berlin, Germany.}
\affiliation{Department of Artificial Intelligence, Korea University, Seoul, South Korea.}
\affiliation{MPI for Informatics, Saarbr\"ucken, Germany.}
\author{Stefan Chmiela}
\thanks{Corresponding authors.}
\affiliation{BIFOLD, Berlin, Germany.}
\affiliation{Machine Learning Group, TU Berlin, Berlin, Germany.}

\input{chapters/0_abstract}
\maketitle
\setcounter{page}{1}
\renewcommand{\thepage}{\arabic{page}}
\raggedbottom 
\input{chapters/1_introduction}
\input{chapters/2_results}

\input{chapters/4_discussion}
\input{chapters/6_methods}
\input{chapters/acks_decls}

\pagebreak

\bibliography{refs}

\input{chapters/9_appendix}

\end{document}

%% file: preamble.tex
\usepackage{amsmath}
\usepackage{booktabs}
\usepackage{graphicx}
\usepackage{dcolumn}
\usepackage{bm}
\usepackage{xfrac} 
\usepackage[HTML]{xcolor} 
\usepackage{mathtools} 
\usepackage{amssymb}
\usepackage{multirow}
\usepackage[normalem]{ulem}
\usepackage{svg}
\usepackage{textcomp}
\usepackage{makecell}
\usepackage{romannum}
\usepackage{siunitx} 
\usepackage{float} 
\usepackage{placeins}
\usepackage{capt-of}

\DeclareSIUnit{\kcal}{kcal}
\DeclareSIUnit{\byte}{B}
\DeclareSIUnit{\angstrom}{\text{\AA}} 
\newcommand{\SIwn}[1]{\SI[per-mode=power]{#1}{\per\centi\meter}} 

\usepackage{tikz}
\usetikzlibrary{arrows.meta, bending}

\usepackage[version=4]{mhchem} 
\usepackage{xspace} 
\usepackage{etoolbox} 

\usepackage{algpseudocode} 

\algnewcommand\StatefulVars{\textbf{State:}}
\algnewcommand\Stateful{\item[\StatefulVars]}

\newcommand{\omitt}[1]{} 
\newcommand{\maybe}[1]{\textcolor{gray}{#1}} 
\newcommand{\keepit}[1]{{#1}} 

\newcommand{\jm}[1]{\ifmmode\textcolor{teal}{\ensuremath{#1}}\else\textcolor{teal}{#1}\fi}  
\newcommand{\jms}[2]{\ifmmode{\ensuremath{\sout{#1} \textcolor{teal}{#2}}}\else{\sout{#1} \textcolor{teal}{#2}}\fi}  

\definecolor{mydarkred}{HTML}{7a2828}
\definecolor{mydarkblue}{HTML}{2c287a}
\definecolor{mycyan}{HTML}{47857a}

\newcommand{\basecommand}[2]{\ifmmode \mathbf{#1}#2 \else \ensuremath{\mathbf{#1}#2}\kern0.em \fi \xspace}
\NewDocumentCommand{\IndexedCmd}{m g}{%
  \basecommand{#1}{\IfNoValueTF{#2}{}{^{(#2)}}}%
}

\newcommand{\hs}[1][]{\basecommand{h}{^*#1}} 
\newcommand{\vs}[1][]{\basecommand{u}{^*#1}} 
\newcommand{\vbars}[1][]{\basecommand{\bar{u}}{^*}} 

\NewDocumentCommand{\h}{g}{\IndexedCmd{h}{#1}}          
\RenewDocumentCommand{\v}{g}{\IndexedCmd{u}{#1}}        
\NewDocumentCommand{\vbar}{g}{\IndexedCmd{\bar{u}}{#1}} 
\NewDocumentCommand{\sq}{g}{\IndexedCmd{\square}{#1}}    

\newcommand{\x}{\basecommand{x}{}} 
\newcommand{\R}{\basecommand{R}{}} 
\newcommand{\M}{\basecommand{M}{}} 

\newcommand{\Z}{\basecommand{Z}{}} 
\newcommand{\F}{\basecommand{F}{}} 
\newcommand{\w}{\basecommand{w}{}} 

\newcommand{\att}[1][0]{_{\attlong{#1}}}
\newcommand{\attlong}[1]{{\left(t\ifstrequal{#1}{0}{}{ #1 \Delta t}\right)}}
\newcommand{\attshort}[1]{{\ifstrequal{#1}{0}{}{\left(#1 \Delta t\right)}}}

\newcommand{\jac}{\textit{jac}\xspace}
\newcommand{\itc}{\textit{itc}\xspace}
\newcommand{\trunc}{\textit{trunc}\xspace}

\newsavebox{\constWarmstartArrow}
\savebox{\constWarmstartArrow}{%
  \tikz[baseline, scale=0.4, transform shape]{
    \draw[dash pattern=on 2pt off 1pt, line width=0.8pt, color=mydarkred,
          -{Stealth[length=4pt, width=2.5pt]}] (0,-0.7) -- (0,0);
  }%
}
\newsavebox{\linearWarmstartArrow}
\savebox{\linearWarmstartArrow}{%
  \tikz[baseline, scale=0.4, transform shape]{
    \draw[dash pattern=on 2pt off 1pt, line width=0.8pt, color=mydarkred, 
    -{Stealth[length=4pt, width=2.5pt]}] (-0.5,-0.7) -- (0,0);
  }%
}
\newsavebox{\twolinearWarmstartArrows}
\savebox{\twolinearWarmstartArrows}{%
  \tikz[baseline, scale=0.4, transform shape]{
    \draw[dash pattern=on 2pt off 1pt, line width=0.8pt, color=mydarkred, -{Stealth[length=4pt, width=2.5pt]}] (0,0) -- (0.7,-0.7);
    \draw[dash pattern=on 2pt off 1pt, line width=0.8pt, color=mydarkred, -{Stealth[length=4pt, width=2.5pt]}] (0.45,0) -- (1.15,-0.7);
  }%
}
\newsavebox{\fixedpointArrow}
\savebox{\fixedpointArrow}{%
  \tikz[baseline, scale=0.4, transform shape]{
    \draw[solid, line width=0.8pt, color=black,
    -{Triangle[length=4pt, width=2.5pt, fill=white, line width=0.5pt]}] (-0.5,-0.7) .. controls (-0.1,-0.25) .. (0,0);
  }%
}
\definecolor{sketchSlateTeal}{HTML}{4A7C7C}
\definecolor{sketchTealGreen}{HTML}{4FAA90}
\definecolor{sketchLimeGreen}{HTML}{9DC840}
\newsavebox{\sketchLinearArrow}
\savebox{\sketchLinearArrow}{%
  \tikz[baseline, scale=0.4, transform shape]{
    \draw[solid, line width=1.8pt, color=sketchSlateTeal, line cap=round,
    -{Stealth[length=5pt, width=4pt, fill=sketchSlateTeal, line width=0pt]}]
      (-0.7,-0.7) -- (0,0);
  }%
}
\newsavebox{\sketchQuadraticArrow}
\savebox{\sketchQuadraticArrow}{%
  \tikz[baseline, scale=0.4, transform shape]{
    \draw[solid, line width=1.8pt, color=sketchTealGreen, line cap=round,
    -{Stealth[length=5pt, width=4pt, fill=sketchTealGreen, line width=0pt]}]
      (-0.7,-0.7) .. controls (-0.85,-0.2) .. (0,0);
  }%
}
\newsavebox{\sketchCubicArrow}
\savebox{\sketchCubicArrow}{%
  \tikz[baseline, scale=0.4, transform shape]{
    \draw[solid, line width=1.8pt, color=sketchLimeGreen, line cap=round,
    -{Stealth[length=5pt, width=4pt, fill=sketchLimeGreen, line width=0pt]}]
      (-0.7,-0.7) .. controls (-1.05,-0.25) and (-0.45,0.1) .. (0,0);
  }%
}

%% file: chapters/0_abstract.tex
\begin{abstract}
We introduce {\em implicit} machine learning force fields (I-MLFFs), which replace {\em explicit} stacks of neural network layers with self-consistent fixed-point equations.
In molecular simulations, this formulation enables intermediate representations to be reused across successive timesteps, thereby warm-starting force evaluation. 
The resulting models effectively combine the computational footprint of a shallow, single-layer MLFF with the representational capacity and accuracy of a deep neural network.
Our approach unlocks architecture-agnostic efficiency gains that are inaccessible when force prediction and trajectory integration are considered separately.
We demonstrate this across three major classes of graph neural networks: invariant, equivariant Cartesian tensor, and SO(3)-equivariant spherical-tensor architectures. 
Each yields a two- to five-fold reduction in compute and memory footprint.
Crucially, these gains are achieved while retaining full atomistic resolution and the original integration timestep, avoiding spatial or temporal coarse graining.
Our contribution therefore advances the scaling frontier of quantum-mechanically faithful molecular simulation, enabling longer trajectories and larger atomistic systems within fixed GPU memory and compute budgets, and thereby opening access to new insights across biomolecular and material systems.
\end{abstract}

%% file: chapters/1_introduction.tex
\section{Introduction}

\begin{figure*}
  \centering
    \includegraphics[width=\linewidth]{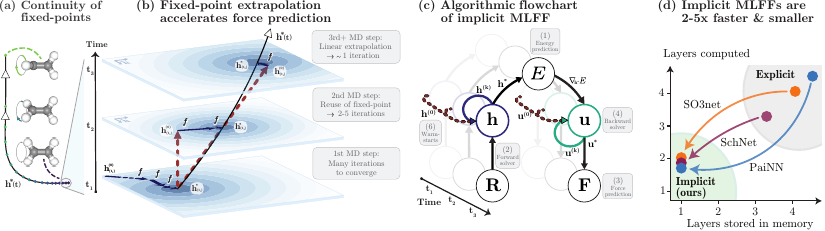}

\caption{
\textbf{Temporal continuity in MD enables computational reuse in implicit MLFFs.}
(a) Learned fixed-point representations $\hs(t)$ of the implicit MLFF evolve smoothly in representation space along a $\mathrm{C_2H_4}$ torsion coordinate (black arrow \raisebox{1.9ex}{\usebox{\fixedpointArrow}}), with larger changes near bond breaking events.
(b) This temporal continuity enables reuse and extrapolation of previous fixed points (red dashed arrows
\raisebox{1.7ex}{\usebox{\constWarmstartArrow}},%
\raisebox{1.7ex}{\usebox{\linearWarmstartArrow}}%
), reducing the number of solver iterations required in subsequent MD steps. The first step (bottom) requires many applications of $f$ to converge, whereas later steps can be warm-started from previous fixed points using fixed-point reuse (middle) or linear extrapolation (top).
(c) Algorithmic flowchart of an implicit MLFF. 
A forward fixed-point solve $\hs = f(\hs, \R)$ (\textcolor[HTML]{2c287a}{\rotatebox[origin=c]{45}{\boldmath$\boldsymbol{\curvearrowleft}$}}) yields $E = f_E(\hs)$. Forces, $\F = -\nabla_{\R} E$, are computed in the backward pass by solving a second fixed-point problem for $\vs$ (\textcolor[HTML]{47857a}{\rotatebox[origin=c]{135}{\boldmath$\boldsymbol{\curvearrowright}$}}). Both solves are warm-started from preceding MD steps. Corresponding equation numbers are indicated. (d)
The computational footprint of implicit models is 2--5$\times$ smaller than explicit models with matching force accuracy, averaged across MD17 and MD22 datasets at \qty{300}{\kelvin}.
Implicit models 
using linear extrapolation need only
1--2 layer evaluations to predict forces, with implicit differentiation requiring only one layer in memory.
In contrast, explicit models have to evaluate, store, and differentiate all their 3--5 layers in every MD step.
}    
\label{fig:intro}
  \end{figure*}

A major obstacle in accurate molecular dynamics (MD) simulations~\cite{Karplus2002MD} of physical systems, such as large proteins in aqueous environments, is the need for millions of computationally expensive quantum mechanical calculations to compute observables that may then be directly compared with experimental results.
Over the past decade, machine learning force field (MLFF) development has been oriented toward this application. Modern MLFFs are several orders of magnitude faster than \emph{ab-initio} methods~\cite{unke2021machine}, while maintaining the accuracy of large-basis set density functional theory (DFT)~\cite{gasteiger2020directional, schutt2021equivariant, ko2021fourth, gasteiger2021gemnet, liao2022equiformer, wang2022visnet, batatia2022mace, batzner20223, musaelian2023learning, frank2024euclidean} and coupled-cluster reference calculations~\cite{chmiela2018towards} for well-defined inference tasks.
Remarkably, the fastest MLFFs available today achieve linear scaling~\cite{kabylda2025molecular}.
While these models can achieve up to $10^6$ simulation steps per day~\cite{kovacs2023mace}, they have still not reached the necessary efficiency for studies at desired practical scales. Many experimentally relevant phenomena, such as protein folding, occur over timescales that span microseconds to milliseconds, requiring billions to trillions of molecular dynamics (MD) simulation steps~\cite{tuckerman2002ab}.
Consequently, classical FFs~\cite{Cornell1995AMBER, MacKerell1998CHARMM} or coarse graining~\cite{wang2019machine, husic2020cgschnet,majewski2023machine, charron2025navigating, durumeric2026learning} remain the methods of choice for studying phenomena at the largest scales, even when the high accuracy of MLFFs would be preferable.
Although continued advances in MLFF architectures and implementations will reduce the cost of individual force evaluations, conventional inference still repeats the full computation at every integration step. We therefore shift the focus from optimizing the MLFF in isolation to exploiting efficiencies across the MD simulation loop as a whole.

In atomistic simulations, the classical Newtonian equations of motion are solved numerically to evolve nuclear positions over time. 
The computational bottleneck arises from the inherent need for fine temporal resolution, typically on the order of one femtosecond ($10^{-15}$ s), to capture high-frequency dynamics of light atoms accurately.
This constraint reflects the Nyquist–Shannon sampling limit. Extending the timestep can lead to instabilities and unrealistic system behavior in the simulation~\cite{lane2013milliseconds}.
However, at room temperature ($\sim$\qty{300}{\kelvin}), the change in atomic positions during a single simulation step is typically on the order of only \qtyrange{0.001}{0.01}{\angstrom} (less than ${\sim}\qty{1}{\percent}$ of a typical covalent bond length in organic molecules), leaving the global structure of the system mostly unchanged.
As a result, successive evaluations of the MLFF along a simulation trajectory are highly similar.
This calls for novel approaches that exploit this inherent redundancy to avoid starting the calculations from scratch at every step.

To address this challenge, we propose using the principle of \emph{implicit} modeling~\cite{simard1988fixed, miller2018stable, bai2019deq, winston2020monotone} to amortize inference across simulation time by systematically building on prior computations.
Implicit models treat inference as the solution of an underlying dynamical system, effectively a continuous-time counterpart to traditional fixed-depth neural network (NN) architectures.
Deep Equilibrium Models~\cite{bai2019deq} are a specific formulation of implicit models in which inference is formulated as a convergent, self-consistent procedure whose limit corresponds to an equilibrium (fixed point) of a differential equation defined by a single learned layer~\cite{lu2018beyond, chen2018neural, haber2017stable, ruthotto2020deep}.
In our setting this equilibrium defines a representation of the molecular graph  from which energies and atomic forces are predicted.
Because the solution is defined as a limit rather than a fixed sequence of computations, the solver can be initialized from arbitrary states.
We use this flexibility to exploit the temporal coherence of simulation trajectories: intermediates from prior steps provide effective initializations for subsequent solves. This enables warm-started inference, substantially reducing the number of iterations needed to converge the latent molecular graph representation. 
Once the fixed point is reached, we further {\em implicitly} differentiate the energy, avoiding the need to unroll the solver to obtain conservative forces.
Together, this reduces the inference cost per simulation step significantly, below that of traditional explicit MLFFs, 
bringing them closer to the practical applicability of classical FFs 
without compromising their characteristic accuracy.

These
gains are achieved
without relaxing any aspect of physical accuracy, which sets apart our approach from many existing strategies for extending accessible simulation timescales, including prior work that exploits the temporal continuity of MD simulations~\cite{schaaf2024boostmd, burger2025dequify}.
Coarse-grained methods, for example, accelerate simulations by sacrificing atomistic resolution~\citep{wang2019machine, husic2020cgschnet,majewski2023machine, charron2025navigating, durumeric2026learning}. 
Other approaches predict molecular evolution by taking large effective time steps~\cite{thiemann2025trajcast, bigi2026flashmd, ripken2026hfm}, which can skip over high-frequency dynamics, such as bond vibrations or rapid hydrogen-bond rearrangements.
A unique aspect of our solution is that it preserves energy conservation, a fundamental property of isolated classical and quantum mechanical systems~\cite{chmiela2017machine}.
In contrast, direct force-prediction methods~\cite{hu2021forcenet, zitnick2022scn, gasteiger2022gemnetoc, passaro2023escn, liao2023equiformerv2, neumann2024orb, eissler2026simple} can introduce non-conservative force components that contribute to thermodynamic drift and runaway instabilities in molecular simulations~\cite{fu2022forces, bigi2024dark}. This limits them to short simulation times that are insufficient for realistic MD applications.
While small energy-conservation violations can be corrected with thermostats, they disrupt the sampling efficiency of the trajectory~\cite{bigi2024dark}. 

%% file: chapters/2_results.tex
\section{Results}
\label{sec:results}

\input{chapters/2a_results_theory}

\input{chapters/2b_results_understanding}
\input{chapters/2c_results_warmstarts}

\input{chapters/2d_results_md}

\input{chapters/2e_iteration_depth}

%% file: chapters/2a_results_theory.tex
Many state-of-the-art MLFFs are built on Graph Neural Networks (GNNs), which encode the molecular geometry into latent atomic representations by iteratively passing
messages along interatomic edges~\cite{unke2021machine,keith2021combining}. Most explicit GNN-based MLFFs can be reformulated as implicit models with only minimal architectural changes.
To demonstrate the general and architecture-independent applicability of our approach, we apply it to three GNN architectures that are considered representative of the main types of MLFFs: the invariant continuous-filter convolutional SchNet architecture~\cite{schutt2018schnet},
the Cartesian tensor message-passing PaiNN architecture~\cite{schutt2021equivariant}, and a generic SO(3)-equivariant spherical-tensor
MPNN, representing scalar, vectorial, and higher-order tensor feature spaces,
respectively.
The latter serves as a typical template for a broad class of contemporary equivariant MLFF~\cite{thomas2018tensor,unke2021spookynet, batzner20223, batatia2022mace, liao2022equiformer, musaelian2023learning, passaro2023escn, liao2023equiformerv2, frank2024euclidean, fu2025esen, wood2025uma}.
GNNs encode the molecular graph $\x := [\R, \Z]$, consisting of the nuclei positions $\R \in \mathbb{R}^{N \times 3}$ and atomic numbers $\Z \in \mathbb{N}_+^N$ of $N$ atoms, into a latent representation $\hs(\x) \in \mathbb{R}^{N {\times} F}$.
From this a readout function $f_E$ infers the energy and forces:
\begin{equation}
E = f_E(\hs) \text{ and } \F = -\nabla_{\R}E.
\label{eq:results_E_F}
\end{equation}
Explicit models form the latent representation of the molecule by composing
a stack of $K$ interaction layers
$
\h{K} = f^{(K)} \circ \dots \circ f^{(1)}(\x)
$, where each layer may have its own learned parameters.
Notably, they offer no mechanism for adaptive compute or warm-starting, and all $K$
layers need to be kept in memory in order 
to compute \F with backpropagation.
By contrast, implicit models iterate
a single, learned interaction layer
$\h{k+1} = f(\h{k}, \x)$
until 
the residual change in \h is small,
reaching a fixed point:
\begin{equation}
\hs = f(\hs, \x).
\label{eq:results_h_star}
\end{equation}
To guarantee the existence of a fixed point, $f$ must map a compact convex set to itself~\cite{Brouwer1911}, which is achieved through normalization.
We further inject \R and \Z at every iteration to ensure that the fixed point remains conditioned on the molecular structure (Sec.~\ref{sec:mlff_modifications}).
By the Implicit Function Theorem (IFT)~\cite{krantz2002implicit}, equation~\eqref{eq:results_h_star} locally defines a differentiable mapping $\x \mapsto \hs(\x)$. This enables efficient force computation by implicit differentiation (cf.~Sec.~\ref{subsec:methods_implicit_forces}), expressing \F entirely in terms of derivatives evaluated at \hs:
\begin{equation}
\F
= - \underbrace{
\left(\frac{\partial f_E}{\partial \hs}\right)^\top 
\left(\mathbb{I} - \frac{\partial f}{\partial \hs}\right)^{-1}
}_{(\vs)^{\top}}
\frac{\partial f}{\partial \R}.
\label{eq:results_force_readout_derivative}
\end{equation}
Because the gradients depend only on \hs, and not on the path taken to reach this fixed point, the forward solver can be chosen freely and does not require unrolling. Only the final evaluation of $f$ must be retained in memory, reducing memory usage by up to a factor of $K$ relative to an explicit $K$-layer network (cf. SI Fig.~\ref{fig:si-memory-lineplot}). This allows larger systems to be simulated on the same fixed-memory GPU.
To avoid explicit matrix inversion, we reformulate the \emph{fixed-point adjoint} \vs in equation~\eqref{eq:results_force_readout_derivative} as the solution of a second fixed-point equation (Sec.~\ref{subsec:methods_implicit_forces}):
\begin{equation}
\vs = 
\left(\frac{\partial f}{\partial \hs}\right)^\top
\vs
+ \frac{\partial f_E}{\partial \hs}.
\label{eq:results_v_star}
\end{equation}
We monitor convergence atom-wise using the residual
$\max_{a \in \text{atoms}} \| \sq{k-1}_a - \sq{k}_a \|_2 / \| \sq{k-1}_a \|_2$ for \h and \v,
and find that a threshold of $10^{-2}$ is sufficient for accurate force prediction (Sec.~\ref{subsec:understanding}).
Fast convergence of the fixed-point iteration is encouraged by a
regularization scheme during training (Sec.~\ref{sec:model_training}) and
further accelerated in MD simulations 
by warmstarting the fixed-point solver with initial guesses
$\h{0} \approx \hs$
and
$\v{0} \approx \vs$
(which we develop in
Sec.~\ref{sec:results-warmstarts}).
The full algorithm for implicit energy-conservative FFs is illustrated in Fig.~\ref{fig:intro}C and written out in Algorithm 1 (SI Sec.~\ref{sec:algorithm}).
Further intuition is provided in SI Sec.~\ref{sec:si_ising} through the mean-field Curie–Weiss Ising model, which reduces to a simple scalar implicit equation that is easier to interpret.

%% file: chapters/2b_results_understanding.tex
\begin{figure*}[!t]
  \centering
  \includegraphics*[width=\linewidth]{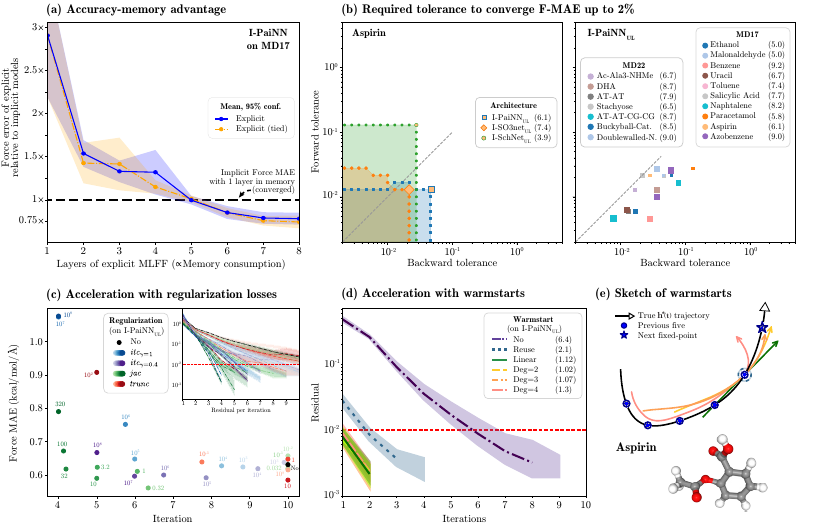}
  \caption{\textbf{Understanding and implementing implicit MLFFs.} 
  (a)~%
  Implicit MLFFs with one layer match or exceed the accuracy of explicit models with up to four layers on MD17.
  We report average force MAEs of 
  I-PaiNN and show the mean with \qty{95}{\percent} confidence interval of the relative error increase $\mathrm{(E-I)/I}$ between explicit $\mathrm{(E)}$ and implicit $\mathrm{(I)}$ models.
  (b)~%
  Implicit models remain accurate with approximate solver convergence.
  Across three architectures (left) and 16 systems (right subplot),
  forward and backward solves require comparable tolerances of at most ${\sim}10^{-2}$ to converge the force error up to \qty{2}{\percent}.
  Encouragingly, the necessary avg. iteration count (shown in brackets in the legends and in the marker size) does not generally increase with the molecule size.
  (c)~%
  Regularization can accelerate convergence without sacrificing accuracy. All four regularization losses reduce iteration counts, but only \itc ($\gamma{=}0.4$) and \jac
  preserve force accuracy.
  We show test MAEs versus the average iterations required to reach a $10^{-2}$ residual, with the inset indicating convergence to this threshold (red line).
  (d)~%
  Warmstarts yield near one-layer implicit MLFFs. Linear extrapolation (equation~\eqref{eq:results_h0}) accelerates convergence by $6\times$ over no warmstart and $2\times$ over fixed-point reuse, while preserving high accuracy. High-order extrapolation can further reduce iteration counts but may overfit.
  (e)~%
  Warmstart construction. Adams--Bashforth-inspired warmstarts extrapolate from the current fixed point using finite differences of previous fixed points to estimate derivatives.
  }
  \label{fig:results-1}
\end{figure*}

\subsection{Implicit MLFFs: Accuracy \& Regularization}
\label{subsec:understanding}

Implicit MLFFs are governed by a trade-off between accuracy and cost: forces are only accurate once the latent fixed point is sufficiently converged, so accuracy is inherently coupled to the number of solver iterations required per prediction. In this section we study this trade-off and ask whether training-time regularization reduces solver cost to the low, MD-compatible budgets that would make implicit models competitive.
%
We begin by evaluating the accuracy of the implicit approach when fully converged. Implicit models reuse one parameterized layer across iterations, whereas explicit models grow parametric capacity with depth, which can potentially translate into higher accuracy. This leads to the question of whether parameter sharing fundamentally limits the implicit approach.
To answer it, we compare a fully converged implicit model to explicit models with 1--8 layers on the MD17 benchmark~\cite{chmiela2017machine} (Fig.~\ref{fig:results-1}A), using PaiNN as the backbone architecture (see Sec.~\ref{sec:model_training} for training details). We also include a parameter-tied explicit variant (sharing parameters across layers) to decouple depth from parameter count in our analysis. 
Remarkably, the prediction performance of the implicit model exceeds both explicit baselines up to four layers, a commonly used depth for MLFFs 
\cite{
schutt2017schnet,
schutt2018schnet,
batzner20223,
batatia2022mace,
grisafi2022equivariant,esders2025analyzing}.
At greater depths, the explicit models do eventually achieve higher accuracies, suggesting that the additional structure and constraints introduced to make the implicit formulation well-defined may act as a limiting factor. Nevertheless, the strong performance at practically relevant depths makes implicit models a parameter-efficient alternative that preserves accuracy where it matters most.

However, fully converging implicit models to machine precision, as in the previous test, is impractical: the required iterations can exceed explicit depth and erase runtime gains. Implicit models are only useful if they yield accurate representations in a few iterations.
Therefore, our second experiment quantifies the convergence tolerances $\epsilon$ needed for the latent fixed-point embedding $\mathbf{h}^*$ to reach a 
target force-prediction accuracy (within $\pm$\qty{2}{\percent} of the fully converged implicit F-MAE).
We compare performance on the MD17-aspirin dataset across three representative MLFF architectures, SchNet, PaiNN, and SO3net (Fig.~\ref{fig:results-1}B).
A key focus is the implicit backward pass used to compute forces efficiently (equation~\eqref{eq:results_v_star}), since the IFT relies on a sufficiently converged fixed point. If the forward iteration is stopped too early, gradients may become inconsistent and any computational savings from early stopping could be offset.
In practice, we observe that sensitivity is approximately symmetric with respect to the forward and backward tolerances.
Restricting both to the same value therefore yields efficient and accurate forces across all three architectures, a simplification that reduces the search space without sacrificing performance.
Encouragingly, all models reach the target force accuracy at comparatively loose convergence thresholds, with tolerances in the range $10^{-2}$--$10^{-1}$.
These results suggest that early stopping is feasible across implicit formulations of different architectures and largely insensitive to parameter settings.
%
Whether a single acceleration technique can transfer broadly depends on how consistently these convergence characteristics hold across systems. To assess this, we repeat the test across molecules spanning different sizes, compositions, and conformational flexibilities, from the MD17~\cite{chmiela2017machine} and MD22~\cite{chmiela2023} benchmarks. Here, we focus on our implicit PaiNN variant (I-PaiNN), 
the more efficient of the two equivariant networks tested
in the previous experiment (Fig.~\ref{fig:results-1}B-II).
We find that forward and backward convergence criteria remain closely aligned across diverse systems. Convergence tolerance is weakly correlated with system flexibility (with the exception of naphthalene and toluene): small, comparatively rigid aromatic molecules (e.g., salicylic acid, paracetamol, aspirin) tend to lie toward the upper right (looser tolerances), while multi-torsional and chain-like systems (e.g., Ac-Ala3-NHMe, stachyose, DHA) cluster somewhat lower.
Notably, we find no evidence that larger molecules require tighter convergence thresholds. For example, the longer DNA segment containing two Adenine--Thymine and two Cytosine--Guanine base pairs (AT-AT-CG-CG) converges in fewer iterations than the shorter DNA duplex segment AT-AT. These results suggest that the implicit approach remains robust to system complexity.

Given that early stopping is feasible and consistent, we next ask whether training-time regularization can push solver cost down to the low iteration counts MD requires. Because fixed points are easier to compute when the forward dynamics are simple, we consider three regularizing losses to improve the stability of our implicit models~\cite{kawaguchi2021theory} and identify which one is the most effective:
Jacobian regularization (\jac)~\cite{bai2021stabilizing} adds a stochastically estimated penalty on the Jacobian of $f$, effectively shrinking its spectral radius and encouraging contraction around the fixed point.
Iterate correction (\itc) explicitly pulls all earlier solver iterations \h{k} toward \hs, encouraging them to converge quickly.
Finally, in truncated prediction (\trunc), energies and forces are derived from the unrolled computation path of early representations \h{1}, \h{2}.
The latter two regularization techniques were adapted from the approach introduced in Ref.~\cite{bai2022optical} (see Sec.~\ref{sec:model_training}).
Fig.~\ref{fig:results-1}C quantifies the resulting trade-off between convergence speed (average number of iterations to reduce the embedding residual below $10^{-2}$, on diverse conformations, cf. SI Sec. \ref{sec:experiment_iter_count_details}) and force accuracy (F-MAE).
The unregularized implicit model (black dot) requires $10$ iterations to reach the residual threshold, yielding an F-MAE of ${\sim}\qty{0.62}{\kcal\per\mole\per\angstrom}$ on the aspirin system.
Increasing \trunc strength accelerates convergence, but systematically degrades force accuracy.
Applying \itc with $\gamma{=}1$ yields even larger speed-ups, yet strong contraction penalties over-regularize the dynamics and lead to substantially less accurate models.
The variant
\itc with $\gamma{=}0.4$ focuses the convergence penalty on late iterations, preserving accuracy.
We find that \jac provides the most favorable Pareto trade-off: it reduces the required iterations from $10$ to ${\sim} 6$ while preserving accuracy (${\sim}\qty{0.6}{\kcal\per\mole\per\angstrom}$ at a regularization of strength $0.32$).
We speculate that the Jacobian penalty improves local conditioning near the solution, accelerating convergence, while leaving the relative position of the fixed points and the function class represented by the model largely unchanged.
Guided by these insights, models throughout the paper are trained using \jac at a moderate strength of 0.32 to stabilize solver dynamics,
and with \itc($\gamma{=}0.4$) at a mild strength of $10^3$ to guarantee convergence within the 10 iterations allotted during training (Sec.~\ref{sec:model_training}).
Taken together, these experiments show that training-time regularization helps but plateaus: even efficient regularizers require ${\sim}6$ calls of an interaction layer, short of the 
strict compute and accuracy demands required for MD.

%% file: chapters/2c_results_warmstarts.tex
\subsection{Warm-Starting Implicit MLFFs}
\label{sec:results-warmstarts}

To improve this performance plateau, we exploit a source of structure unavailable to generic solvers: the temporal coherence along MD trajectories.
The fixed point $\hs(t)=\hs(\R(t))$ varies smoothly along the trajectory because the atomic positions $\R(t)$ evolve continuously under Newton’s equations and the mapping $\R\mapsto\hs(\R)$ is differentiable by the IFT.
Because MD uses small integration timesteps, 
consecutive simulation states $\hs\att, \hs\att[+], \dots$ are therefore close. 
This motivates the central objective of this work: warm-starting both the fixed-point and adjoint solves to accelerate the force prediction.
In the simplest case, the fixed point from the previous step provides a natural initial guess for the next solve $\h{0}\att[+] {:=} \hs\att$, a zeroth-order (constant) approximation of $\hs(t)$ used by prior work~\cite{bai2022optical, burger2025dequify}. We instead exploit the smoothness of the trajectory more carefully:
to increase the accuracy of the warmstart further, we propose a first-order approximation (i.e., a linear extrapolation) $T_1$ 
across the MD time step $\Delta t$:
\begin{equation}
\hs(t + \Delta t) \approx 
T_1(t + \Delta t) = \;
\hs(t) + \frac{d\hs}{dt}\bigg|_{t} \Delta t.
\end{equation}
We adopt a backward finite difference approximation for the derivative,
$\frac{d\hs}{dt}\big|_{t}\Delta t \approx \hs\att - \hs\att[-]$,
and initialize the fixed-point iteration 
in equation~\eqref{eq:results_h_star}
at $t + \Delta t$
with:
\begin{equation}
\h{0}\att[+] := 2\hs\att - \hs\att[-].
\label{eq:results_h0}
\end{equation}
However, to drive efficient MD with energy-conservative forces, not only the energy prediction in the forward pass but also the backward pass used for force derivation needs to be accelerated.
A key methodological contribution of this work is the analogous warm-starting of the iterative solver used for implicit differentiation in equation~\eqref{eq:results_v_star} (Sec.~\ref{subsec:methods_implicit_forces}).
This linear predictor generalizes to higher-order polynomial extrapolators using additional predecessors, similar to Adams--Bashforth multi-step integrators (SI Sec.~\ref{sec:si-warmstarts-extended}).
We note that higher orders can reduce iterations further but eventually risk overfitting to the local trajectory (Fig.~\ref{fig:results-1}E).
This scheme incurs negligible overhead as it only relies on inexpensive combinations of already computed fixed points (SI Sec.~\ref{sec:si-memory}).

\begin{figure*}[!t]
\centering
  \includegraphics*[width=\textwidth]{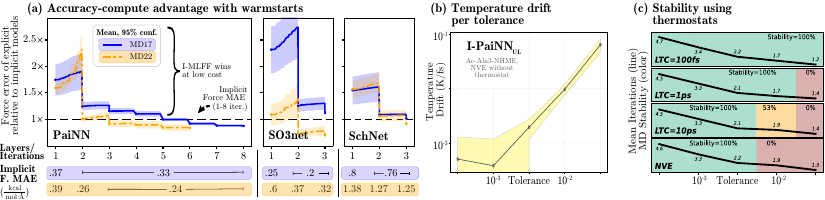}
  \caption{\textbf{Accuracy--cost tradeoffs and stability of implicit MLFFs in MD simulations.}
(a) Implicit MLFFs outperform explicit baselines at matched computational cost, measured by interaction-layer calls averaged over MD17/MD22 trajectories. With linear fixed-point extrapolation from the two previous MD steps and solver tolerance $10^{-2}$, they reach most of their accuracy after one iteration and typically converge within two calls.
We report mean force MAEs over $20{,}000$ test samples and the relative error increase, $\mathrm{(E-I)/I}$, between explicit $\mathrm{(E)}$ and implicit $\mathrm{(I)}$ models at equal average compute.
(b) Energy drift in NVE simulations per solver tolerance.
Plotted is the mean temperature drift with \qty{95}{\percent} confidence intervals over $15 \times \qty{50}{\pico\second}$ MD trajectories initialized at \qty{300}{\kelvin} run in NVE (without a regulating thermostat).
Both (b) \& (c) simulate Ac-Ala3-NHMe using an I-PaiNN$_\textsc{UL}$ model.
(c) Stability of \qty{50}{\pico\second} MD simulations across solver tolerances and thermostat strengths. Background color gives the fraction of stable simulations; black lines and annotations show the mean implicit iteration count. Stronger thermostats suppress energy drift and broaden the stable tolerance range.
}
\label{fig:results-accuracy-stability}
\end{figure*}
Fig.~\ref{fig:results-1}D evaluates the effectiveness
of warmstarting simulations on the precomputed MD17-aspirin trajectory,
averaged over both the forward and adjoint solve,
with six different initializations:
a baseline without warmstarts,
the constant predictor (reusing the predecessor),
the linear predictor (equation~\eqref{eq:results_h0}),
and polynomial predictors of degree $k{=}2$--$4$, using the $k{+}1$ predecessors,
\mbox{$\hs_{(t-\Delta t)}$ to $\hs_{(t-(k+1)\Delta t)}$}.
We solve to the $10^{-2}$ residual threshold established above on a broad set of conformations as described in SI Sec.~\ref{sec:experiment_iter_count_details}.
We find that warm-starting consistently and substantially reduces iteration cost compared with the no-warm-start baseline.
Notably, linear extrapolation converges after only 1.12 iterations on average, a striking six-fold speedup over the no-warm-start reference and an approximate two-fold improvement over simply reusing the preceding solution.
We observe that most conformations converge after a single iteration, effectively reducing inference cost to that of a one-layer neural network. Additional iterations are invoked adaptively only for more challenging configurations, such as those encountered in high-energy regions (cf. Fig.~\ref{fig:results-iteration-depth}).

%% file: chapters/2d_results_md.tex
\subsection{Fast Molecular Dynamics: Accuracy \& Cost}

In the following we will assess whether warmstarts can give implicit models an accuracy--compute advantage against their explicit counterparts in true MD simulations driven by the I-MLFF.
To make the comparison independent of the force-field architecture and implementation, we measure cost as the number of calls to the interaction layer $f$, averaged over the forward and backward pass.
Other contributions, such as the energy-readout head $f_E$, are generally negligible and comparable across all models.
For implicit models, the number of layer calls is controlled by the fixed-point solver tolerance.
Throughout, the solvers are warm-started with an initial guess obtained by linear extrapolation from the two previous steps as in equation~\eqref{eq:results_h0}.
Fig.~\ref{fig:results-accuracy-stability} reports the relative dataset-averaged force MAEs of implicit vs. explicit variants as a function of this compute for PaiNN, SO3net, and SchNet on the MD17 and MD22 datasets.
Implicit models are consistently more accurate in the low-cost regime of $1{-}2$ layer calls, often by a large margin.
At a single layer iteration, I-PaiNN reaches a dataset-averaged force MAE of 0.37/0.39 (MD17/MD22) versus 0.75/0.84 for its explicit counterpart; I-SO3net scores 0.25/0.60 versus 0.76/0.70 (all in \si{\kcal\per\mole\per\angstrom}).
Beyond ${\sim} 3$ layer calls, explicit models become more accurate, but their advantage stays modest (at most ${\sim}\qty{25}{\percent}$) even at large layer counts, where they carry many times the parameters of the implicit model.
Where explicit models first cross over implicit accuracy is architecture- and dataset-dependent: it occurs as early as two layers (PaiNN and SO3net on MD22) but only at five layers for the PaiNN architecture on MD17.

\subsection{Stability of Molecular Dynamics}

To be practically useful in MD, MLFFs must not only be efficient, but also remain stable over long simulation timescales. Accordingly, we next assess the stability of the implicit model on the Ac-Ala3-NHMe molecule while varying the tolerance of both solvers (Fig.~\ref{fig:results-accuracy-stability}B).
Stability depends on the solver tolerance because incomplete convergence in the forward pass violates the assumptions required for implicit differentiation in the backward pass. As a result, the backward pass can yield forces that are inconsistent with the computed energy, which breaks energy conservation and manifests as a spurious temperature drift.
We therefore quantify stability by the mean temperature drift (per \si{\femto\second}) in microcanonical (NVE) trajectories run without a thermostat, so that any systematic drift directly reports non-conservative force inconsistencies. We average over $15$ trajectories of \qty{50}{\pico\second}, each initialized at ${\sim}\qty{300}{\kelvin}$ from a relaxed MD22 conformation (SI Sec.~\ref{sec:experiment_stability_details}).
Observed drifts are close to zero on fine tolerances and show a gradual increase for looser configurations.
In practice, small drifts can be mitigated by running with a Langevin thermostat in the canonical (NVT) ensemble. The thermostat drives the system toward the target temperature on a characteristic Langevin coupling time (LTC).

\begin{figure*}[!t]
  \centering
  \includegraphics[width=\linewidth]{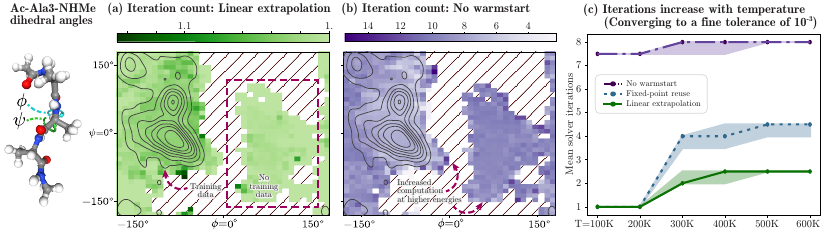}
  \caption{\textbf{Extrapolation behavior and solver dynamics of implicit MLFFs.}
(a,b) Ramachandran plots for Ac-Ala3-NHMe from implicit MLFF (\keepit{I-PaiNN$_\textsc{UL}$}).
Although training conformations are largely confined to $\phi {<} 0^\circ$ (black contours), an MD run at \qty{400}{\kelvin} for \qty{1}{\nano\second} samples an additional free-energy minimum outside this region.
This indicates that implicit models preserve the extrapolative sampling behavior of their explicit counterparts.
Color intensity visualizes mean fixed-point iteration counts in $10^\circ$ $(\phi,\psi)$ bins for warm-started (a) and independently initialized (b) trajectories. Warm-starting keeps the solver close to the one-iteration limit across most conformations, while adaptively increasing iterations in high-energy states that likely have rapid changes in momenta.
(c) Temperature dependence of iteration counts with cold- and warm-started solves \keepit{on I-PaiNN$_\textsc{UL}$}.
Iterations increase with temperature, but warm starts with linear extrapolation reduce this growth.
Plotted are iterations needed to reach a fine tolerance of $10^{-3}$.
Reaching $10^{-2}$ takes ${\sim}1$ iteration using both warmstarts, (SI Fig.~\ref{fig:si-ala3-supplement}).
}
  \label{fig:results-iteration-depth}
\end{figure*}

We validate for a range of solver tolerances which thermostat coupling strength is sufficient to regulate the induced temperature drift and keep the simulation stable (Fig.~\ref{fig:results-accuracy-stability}C). 
Across 15 independent MD trajectories we report the fraction of runs remaining stable by completing the full \qty{50}{\pico\second} horizon without an unphysical bond-breaking event (any bonded interatomic distance exceeding twice its mean in the MD22 reference).
Because overly strong coupling can distort dynamics and potentially bias the integrator, we focus on weak-to-moderate thermostats. 
Under these conditions, the thermostat provides sufficient regulation to offset the drift produced by the implicit model even at a relatively loose tolerance of $10^{-2}$, enabling substantially faster simulations (fewer solver iterations) without compromising practical stability in NVT.
Without a thermostat, NVE runs remain stable up to a tolerance of $3.2 \times 10^{-3}$, averaging $2.2$ solver iterations per step. A moderate coupling of $\mathrm{LTC}{=}\qty{1}{\pico\second}$ lowers this to $1.7$, and a stronger $\mathrm{LTC}{=}\qty{100}{\femto\second}$ to just $1.2$ iterations on average. Again, this brings the computational cost close to the practical lower bound of one solver iteration per step, even in this realistic MD setting.
The cost-matched baseline fails this test: a one-layer explicit model never reaches reliable simulation, passing only \qty{87}{\percent} of runs even at $\mathrm{LTC}{=}\qty{100}{\femto\second}$ and \qty{67}{\percent} in NVE. Implicit models therefore retain stability at near-minimal per-step cost, whereas the explicit counterparts do not.

%% file: chapters/2e_iteration_depth.tex
\subsection{Generalization and Adaptive Iteration Depth}

Next, we examine whether reformulating an MLFF as an implicit model preserves its ability to generalize beyond the training distribution.
To visualize this, we first reproduce a generalization test from Ref.~\cite{frank2024euclidean} on the flexible peptide Ac-Ala3-NHMe:
a \qty{1}{\nano\second} MD simulation with I-PaiNN$_{\textsc{UL}}$ at \qty{400}{\kelvin} correctly explores previously unseen regions of configuration space, consistent with the corresponding explicit model.
The black contours overlaid on the Ramachandran plots in Fig.~\ref{fig:results-iteration-depth}a,b
show the density of the MD22 training set aggregated onto the two backbone angles $\phi, \psi$ of the molecule.
The MD simulation driven by the implicit model discovers the chemically correct minimum~\cite{frank2024euclidean} on the right side of the plot ($\phi{>}0$), which is absent in the training data.

We then use the same MD trajectory to show how \mbox{I-MLFFs} dynamically allocate computation across chemical space, concentrating effort where it is needed most.
Fig.~\ref{fig:results-iteration-depth}a,b visualizes the average iteration count (with and without warmstarts) in $(10^\circ)^2$ bins of the Ramachandran plot.
We find that the simulation becomes highly efficient using warmstarts (Fig.~\ref{fig:results-iteration-depth}a), with only a small fraction of MD steps taking more than one iteration to solve.
Both plots highlight a distinctive advantage of implicit models: the
number of solver iterations adapts for chemically rare or out-of-distribution conformations, such as transition pathways, and converges particularly quickly for frequent conformations near common energy minima, especially when using warmstarts.

Fig.~\ref{fig:results-iteration-depth}c generalizes this finding by showing the effect of temperature on the number of iterations required to reach a residual threshold of $10^{-3}$, chosen here instead of the standard $10^{-2}$ to better resolve differences in iteration count.
We average over $10 \times \qty{500}{\femto\second}$ MD trajectories of Ac-Ala3-NHMe each at \qtyrange{100}{600}{\kelvin} (SI Sec. \ref{sec:experiment_iter_count_details}).
Simulations at temperatures of ${\geq}\qty{300}{\kelvin}$
increase the iteration count of the cold-started solver,
likely because they sample less inside well-known minima
and more in the high-iteration-count regions visible in Fig.~\ref{fig:results-iteration-depth}b.
Warmstarted solvers show a similar rise in iteration count with temperature: at low temperatures, small atomic displacements enable near-perfect amortization and one-step convergence, whereas at higher temperatures the previous fixed point becomes an increasingly poor initial guess. Linear extrapolation mitigates this effect by accounting for the evolution of \hs , thereby maintaining accurate warmstarts and low solver cost.
SI Sec.~\ref{sec:si_adaptive_iteration_depth} provides additional evidence that high potential energy correlates with high iteration counts in the MD of Fig.~\ref{fig:results-iteration-depth}~(a). SI Sec.~\ref{sec:xiv_vibrational_spectra} extends this analysis to another physical observable and shows an increased modeling fidelity of implicit models in high-frequency vibrational spectra of ethanol.
Finally, SI Sec.~\ref{sec:cumulene} examines the generalization properties of implicit models in greater depth on a dataset of cumulene structures \ce{C_nH4} ($n=2,\dots,9$).
Their simple linear carbon-chain geometry provides a controlled setting to show that implicit models can accurately capture long-range electronic effects requiring up to ten message-passing steps, the longest range accessible in this dataset.
Our results indicate that, given a sufficiently strong training signal for long-range interactions, implicit modeling can overcome the finite effective receptive field that limits how far explicit models propagate geometric information.
Our results further show that, when trained only on shorter cumulenes ($n{\leq}6$), implicit models extrapolate accurately to longer chains in both energy and force profiles, unlike explicit models.
Together, these results suggest that implicit depth not only extends the range over which geometric information can be integrated, but also improves extrapolation beyond the molecular sizes seen during training.

%% file: chapters/4_discussion.tex
\section{Discussion \& Conclusion}
\label{sec:discussion}

Standard MLFF inference is stateless: at every femtosecond timestep during an MD simulation, the latent molecular representation needed for force prediction is reconstructed from scratch, even though successive atomic configurations differ only marginally. The implicit MLFF framework introduced here recasts this computation as a self-consistent fixed-point problem, enabling each force evaluation to be warm-started from the preceding solution and thereby amortized along the trajectory. We further improve these initial guesses by polynomially extrapolating previous solutions along the simulation trajectory.
Our results show that a single interaction layer, iterated to convergence, matches the force accuracy of conventional explicit MLFFs that typically use up to five distinct layers and substantially more parameters.
This advance thus directly addresses a key bottleneck in scaling MLFFs to larger systems, where speed and GPU memory are often the limiting constraints.

Beyond atomistic simulation, the implicit modeling framework developed here may provide a route to accelerating forward inference and derivative computations in other physics-based ML applications that rely on iterative simulation.
Examples include neural solvers for quantum-mechanical and electronic-structure problems, such as neural wavefunctions~\cite{spencer2020better, hermann2020deep}, learned or accelerated self-consistent-field methods~\cite{schutt2019unifying, song2026neuralscf, zhang2024self, wang2024infusing}, Lagrangian~\cite{cranmer2020lagrangian} and Hamiltonian neural networks~\cite{greydanus2019hamiltonian} (demonstrated in SI Sec.~\ref{sec:xiii_HNN}),
and fluid models in which pressures, constraints, or energy gradients are obtained through differentiation. Related ideas also appear more broadly in generative modeling: energy-based models, diffusion and score-based models~\cite{sohl2015deep, song2020score, nichol2021improved}, including their probability-flow ODE formulations~\cite{chen2018neural}, as well as continuous normalizing flows~\cite{grathwohl2018ffjord, papamakarios2021normalizing}, all perform inference by integrating learned forces, scores, or vector fields and could thus profit from our framework.

In summary, our results establish temporal continuity in molecular trajectories as a computational resource that can be exploited directly through implicit model design. Across three GNN architectures, two molecular benchmarks, and systems spanning nearly two orders of magnitude in atom count, our I-MLFF framework translates into a consistent two- to five-fold simulation speedup and smaller memory footprint. Because it builds directly on existing GNN architectures, its improvements are orthogonal to (future) architectural advances, implementation optimizations, and improved training-data sampling strategies.

%% file: chapters/6_methods.tex
\section{Methods}

\subsection{Constructing Implicit MLFFs}
\label{sec:mlff_modifications}

In conventional GNN-based (explicit) MLFFs, the molecular representation \h used to predict the potential energy and atomic forces (equation.~\ref{eq:results_E_F}) is constructed by successively refining atom-wise features. Starting from learned embeddings of the atomic types, $\h{0} := \h_{\Z}$, the representation is updated through a fixed sequence of \emph{distinct} interaction layers, $\h{k} = \textsc{Interact}^{(k)}_\R(\h{k-1})$, each of which incorporates geometric information from the nuclear positions.
In contrast, our implicit MLFF approach predicts these outputs from a self-consistent molecular representation, $\hs = f(\hs, \x)$, defined as the fixed point of a \emph{single} learned interaction layer $f$ that depends implicitly on $\x := [\R, \Z]$. To convert an existing explicit MLFF into the implicit form, we introduce two small modifications that ensure the existence of such a fixed point while preserving the general structure of the original architecture:
\begin{equation}
\begin{aligned}
    \h{k}
    =& \; f(\h{k-1}, \x) \\
    :=& \; \textsc{Norm} \circ \textsc{Interact}_{\R}
        \bigl(\h{k-1} + \h_{\Z}\bigr).
\end{aligned}
\end{equation}
First, we add a dependence on $\h_\Z$ into the interaction layer, in order to make the fixed point a function of the atomic numbers, and not only of the atomic positions \R.
This is essential in the implicit setting, where the converged representation no longer retains information from the initial state \(\h{0}\) used to start the solver.
Second, we test two different normalizations
similar to those being used in explicit models to enable depth scaling, to bound the interaction layer outputs.
A simple equivariant implementation of the root mean square norm~\cite{zhang2019rmsnorm} 
rescales \h on each atom $a$ and each of its feature blocks $b \in \{\text{invariant}, \text{equivariant}\}$, 
independently to \textsc{u}nit \textsc{l}ength,
$
\textsc{Norm}_{\textsc{UL}}
= \h_{a,b}
/ (\| \h_{a,b} \|_2 + \eta),
$
with a small positive $\eta = 10^{-5}$ ensuring continuity.
The more expressive equivariant merged layer norm~\cite{liao2026equiformerv3} ($\textsc{Norm}_{\textsc{LN}}$) 
first subtracts the mean from the invariant features,
and then divides all invariant and equivariant features per atom by their \textit{merged} norm.
It then scales each invariant and equivariant feature by a learned factor
and adds a learned offset on each invariant feature.
This increases the force accuracy of implicit models by around \qty{20}{\percent} on most molecules (SI Table~\ref{tab:accuracy}).
The main results in Fig.~\ref{fig:results-1}a and Fig.~\ref{fig:results-accuracy-stability}a report accuracy using $\textsc{Norm}_{\textsc{LN}}$ models; qualitative results using the simpler norm are denoted as e.g. I-PaiNN$_{\textsc{UL}}$.
SI Sec.~\ref{sec:si_accuracy_tables} highlights accuracy differences between the two norms per molecule.
Importantly, both normalizations make $f$ a continuous self-map on a compact space, guaranteeing the existence of a fixed point by Brouwer's fixed-point theorem~\cite{Brouwer1911}.

\subsection{Implicit Differentiation And Force Inference}
\label{subsec:methods_implicit_forces}

At each simulation step, we compute energy-conservative forces by implicitly differentiating the converged fixed point. Because implicit differentiation depends only on local derivative information at this fixed point, it avoids backpropagation through the full solver trajectory. The resulting forces are therefore independent of the warm start used to initialize the solver. Moreover, derivative computation requires retaining only the final application of $f$, so the memory cost matches that of a single-layer explicit model and does not grow with the number of fixed-point iterations used to obtain \hs.

The \textit{implicit} Jacobian $\frac{d \hs}{d \R}$ in equation~\eqref{eq:results_force_readout_derivative} is derived by differentiating the fixed-point equation \mbox{$\hs{(\x)} = f( \hs{(\x)}, \x )$} with respect to the atomic position $\R$ and collecting terms:
\begin{align*}
  &&
  \frac{d \hs}{d \R} 
  &= \frac{\partial f}{\partial \hs}\frac{d \hs}{d \R} 
    + \frac{\partial f}{\partial \R}
  \\
  &\Leftrightarrow&
  \frac{d \hs}{d \R} 
  - \frac{\partial f}{\partial \hs}\frac{d \hs}{d \R} 
  &= \frac{\partial f}{\partial \R}
  \\
  &\Leftrightarrow&
  \left(
        \mathbb{I} - {\frac{\partial f}{\partial \hs}}
  \right)
  \frac{d \hs}{d \R} 
  &= \frac{\partial f}{\partial \R}
  \\
  &\Leftrightarrow&
  \frac{d \hs}{d \R}   
  &= \left(
        \mathbb{I} - {\frac{\partial f}{\partial \hs}}
  \right)^{-1} \frac{\partial f}{\partial \R}
\end{align*}
In the definition of the interatomic forces in equation~\eqref{eq:results_E_F},
the inverse term is absorbed into the \emph{fixed-point adjoint} \vs:
\begin{align*}
-\F
&= \left(\frac{\partial f_E}{\partial \hs}\right)^\top \frac{d \hs}{d \R}
\\
&=
\underbrace{
  \left(\frac{\partial f_E}{\partial \hs}\right)^\top
  \left(\mathbb{I} - \frac{\partial f}{\partial \hs}\right)^{-1}
}_{(\vs)^{\top}}
\frac{\partial f}{\partial \R}
= (\vs)^{\top} \frac{\partial f}{\partial \R}
\end{align*}
We then redefine \vs as a linear self-consistency equation
that can be solved iteratively:
\begin{align*}
&&
\vs
&=
\left(\mathbb{I} - \frac{\partial f}{\partial \hs}\right)^{-\top}
\frac{\partial f_E}{\partial \hs}
\nonumber \\
&\Leftrightarrow&
\left(\mathbb{I} - \frac{\partial f}{\partial \hs}\right)^{\top}
\vs
&=
\frac{\partial f_E}{\partial \hs}
\nonumber \\
&\Leftrightarrow&
\vs  - \left(\frac{\partial f}{\partial \hs}\right)^{\top} \vs
&=
\frac{\partial f_E}{\partial \hs}
\nonumber \\
&\Leftrightarrow&
\vs
&= \left(\frac{\partial f}{\partial \hs}\right)^{\top} \vs
+ \frac{\partial f_E}{\partial \hs}
\end{align*}

\subsection{Training Implicit MLFFs}
\label{sec:model_training}

We train all implicit models without warmstarts, using $\h{0}=\h_{\Z}$ as initialization and
unrolling ten iterations of the computation graph to $\h{10}$.
We find that this method efficiently approximates a converged fixed point if the residual between
\h{10} and \h{9} is smaller than 0.01.
Energy and force predictions are trained jointly with a mean-squared-error (MSE) loss over the $n$ atoms of each system, weighting the force contribution by $\alpha=$\qty{95}{\percent}:
\begin{equation}
  \mathcal{L}_{E,\F}
  = (1-\alpha) \cdot |E - E_\textbf{True}|^2
  + \alpha / n \cdot ||\F - \F_\textbf{True}||_2^2 .
  \label{eq:loss_e_f}
\end{equation}

\paragraph{Regularization losses.}
\label{par:regularization_losses}
We test three additional loss functions to encourage fast convergence of implicit solvers (see Sec. \ref{subsec:understanding}):
Jacobian Regularization (\jac)~\cite{bai2021stabilizing}
promotes a smaller spectral radius
$\rho(\frac{\partial f }{ \partial \hs})$, thereby improving convergence in equations~\eqref{eq:results_h_star} and \eqref{eq:results_v_star}.
In practice, the spectral radius is upper bounded by the Frobenius norm,
which is estimated following Girard~\cite{girard1989},
$$
\mathcal{L}_{\jac} = 
\left\|\epsilon^{\top} \frac{\partial f }{ \partial \h} \bigg|_{\hs} \right\|_2^2,
\text{\;\;with\;\;}
\epsilon \sim \mathcal{N}\left(0, I_d\right).
$$
A grid search on the aspirin dataset in Sec. \ref{subsec:understanding} showed that applying Jacobian regularization with a moderate coefficient reduces iteration counts without compromising force accuracy.
Accordingly, we train all implicit models in this study with $\mathcal{L}_{\jac}$ with a moderate coefficient of 0.32.

Another loss, `truncated prediction' (\trunc), explicitly derives
$E^{(k)} {=} f_E(\h{k})$ and $\F {=} {-} \nabla_{\R}E^{(k)}$
from the early representations \h{1} and \h{2} and minimizes their distance to labels, akin to equation~\eqref{eq:loss_e_f}.
Unlike fixed-point correction~\cite{bai2022optical}, we limit supervision to early iterates -- each iterate's force loss requires second-order backpropagation, adding the compute and memory footprint of $k$ interaction layers per training step.
To avoid this overhead, we propose the lightweight 
`iterate correction' (\itc) 
which does not fit force labels but minimizes the Euclidean distance of all solver iterates to the final representation.
In full,
$
\mathcal{L}_\itc^\gamma
= \sum_k \gamma^{10-k} 
||\h{k}-\h{10}||_2^2.
$
Here, we let gradients propagate only through \h{k}, not \h{10}, and can exponentially downweight the loss on early iterates by setting $\gamma{<}1$.
The grid search (Sec. \ref{subsec:understanding}) shows that
both \trunc and \itc with $\gamma{=}1$ (weighting all iterates equally) reduce iteration counts at the cost of lower predictive accuracy.
Setting $\gamma{=}0.4$ focuses \itc on the convergence of late iterates. 
We apply this variant with a mild loss coefficient of $10^4$ on all following implicit models as it guarantees convergence of \h{k} within the 10 iterations allotted during training without compromising accuracy.
All losses are weighted by their coefficients and summed,
giving
$
\mathcal{L}
= \mathcal{L}_{E,\F}
+ 0.32 \cdot \mathcal{L}_\text{\jac}
+ 10^4 \cdot\mathcal{L}_\text{\itc}^{\gamma=0.4}
$.

\paragraph{Convergence criteria.}
\label{par:hyperparameters}

All models are trained with the AdamW optimizer~\cite{loshchilov2019decoupled} until the loss $\mathcal{L}$ on the validation set shows no improvement for 500 epochs.
The learning rate is initialized to $10^{-3}$ and is halved whenever the validation loss has not improved for 250 epochs.

\subsection{MLFF Architectures}

We use the implementations of SchNet, PaiNN, and SO3net provided by
SchNetPack~\cite{schutt2018schnetpack, schutt2023schnetpack} for training, testing, and MD simulations.
Our extensions comprise input injection and normalization for all three architectures, the addition of the fixed-point solver, implicit differentiation, warm-start logic, and the regularization losses described above.

The three analyzed architectures encode molecular geometry at increasing fidelity.
SchNet acts purely on invariant features and views the geometry of the system only through the pair-wise distances between atoms~\cite{schutt2017schnet}.
PaiNN represents atomic neighborhoods by scalar ($l=0$) and Cartesian equivariant vector ($l=1$) features, while SO3net uses spherical-harmonic features up to degree $L_{\max}{=}2$.
The latter is representative of a broad class of modern equivariant architectures~\cite{liao2022equiformer, wang2022visnet, batatia2022mace, batzner20223, musaelian2023learning, frank2024euclidean}.
All models are configured with a cutoff of \qty{5}{\angstrom}, 50 radial basis functions, and the SiLU activation function~\cite{hendrycks2016gaussian}.

\subsection{Datasets}
\label{sec:datasets}

We train explicit MLFFs of varying depth and implicit models with one interaction layer on each molecule of the MD17 and MD22 datasets.
MD17 contains trajectories of 
10 small molecules containing 9 to 24 atoms each.
Reference energies and forces are calculated with DFT at the pbe+vdw-ts standard,
driving simulations with a timestep of \qty{0.5}{\femto\second}, at a temperature of \qty{500}{\kelvin}~\cite{chmiela2017machine}.
MD22 contains seven biomolecules and supra-molecules ranging from 42 up to 370 atoms.
Each trajectory is sampled at a time resolution of \qty{1}{\femto\second} at temperatures ranging from \qtyrange{400}{500}{\kelvin}; energies and forces are calculated at the pbe+mbd standard~\cite{chmiela2023}.
We use the same random dataset splits across training runs of different models.
On molecules of the MD17 dataset we use 950 training and 50 validation conformations
and increase to 5000 and 1000 on the more complex systems in MD22.
Because of the lower number of available conformations, we reduce training points to 3000 on the `buckyball catcher' and `double-walled nanotube' systems.
Test errors throughout the paper refer to mean absolute force errors of the remaining conformations in \si{\kcal\per\mole\per\angstrom}.

The sequential nature of the MD trajectories in both datasets allows us to measure the precision achieved with warmstarts and a limited compute budget:
Fig.~\ref{fig:results-accuracy-stability} measures this on $20{,}000$ randomly chosen test conformations.
First, their predecessor conformations in the reference trajectory are solved by the implicit model to a tolerance that is realistic in a production MD ($10^{-2}$).
The predecessor fixed points are then used to predict the fixed point of the test conformation, analogously to our method during MD.
Finally, a scan over possible iteration depths gauges the accuracy-speed tradeoff of the warmstarted implicit model.

\subsection{Code Availability}
The source code for training, testing, and running molecular dynamics simulations with I-MLFF can be found at
\url{https://github.com/johannesmaess/imlff}.

%% file: chapters/acks_decls.tex
\section{Acknowledgments}

JM, LW, JTF, WR, MM, KRM, and SC acknowledge support by the German Federal Ministry of Research, Technology and Space (BMFTR) under Grants BIFOLD24B, BIFOLD25B, 01IS18037A, 01IS18025A, and 01IS24087C.
KRM was partly supported by the Institute of Information \& Communications Technology Planning \& Evaluation (IITP) grants funded by the Korea government (MSIT) (No.2019-0-00079, Artificial Intelligence Graduate School Program, Korea University and No. 2022-0-00984, Development of Artificial Intelligence Technology for Personalized Plug-and-Play Explanation and Verification of Explanation) and also by DFG and HFA. KRM  was furthermore supported in parts by the Korea University Grant.
Correspondence to KRM and SC.
We thank Julia Henkel, Alin Banka, Tim Ebert, and Gregor Lied for continuous feedback on our research.
JM thanks Ana Runjić, Saskia Öztürk, Rebekka Maeß, Aslıhan Yüksel, Nicole Trappe, and Jan Vincent Szlang for support, design, and proofreading of the manuscript.

%% file: chapters/9_appendix.tex
\clearpage
\onecolumngrid 
\appendix

\setcounter{page}{1}
\setcounter{figure}{0}
\setcounter{table}{0}

\renewcommand{\thepage}{S\arabic{page}}
\renewcommand{\thesection}{S\arabic{section}}
\renewcommand{\thetable}{S\arabic{table}}
\renewcommand{\thefigure}{S\arabic{figure}}

\input{chapters/9a_cumulene_generalization}
\input{chapters/9b_Spectra}

\input{chapters/9c_streamplot_details}
\input{chapters/9d_energy_to_iteration}

\input{chapters/9e_accuracy_tables}
\input{chapters/9f_memory_plot}

\input{chapters/9g_algorithm}
\input{chapters/9h_warmstarts_ab}

\input{chapters/9i_iteration_increases}

\input{chapters/9j_ising}
\input{chapters/9k_HNN}

%% file: chapters/9a_cumulene_generalization.tex
\section{Long-range effects and generalization in cumulene molecules}
\label{sec:cumulene}

\begin{figure*}[h!]
    \centering
    \includegraphics[width=\linewidth,height=0.85\textheight,keepaspectratio]{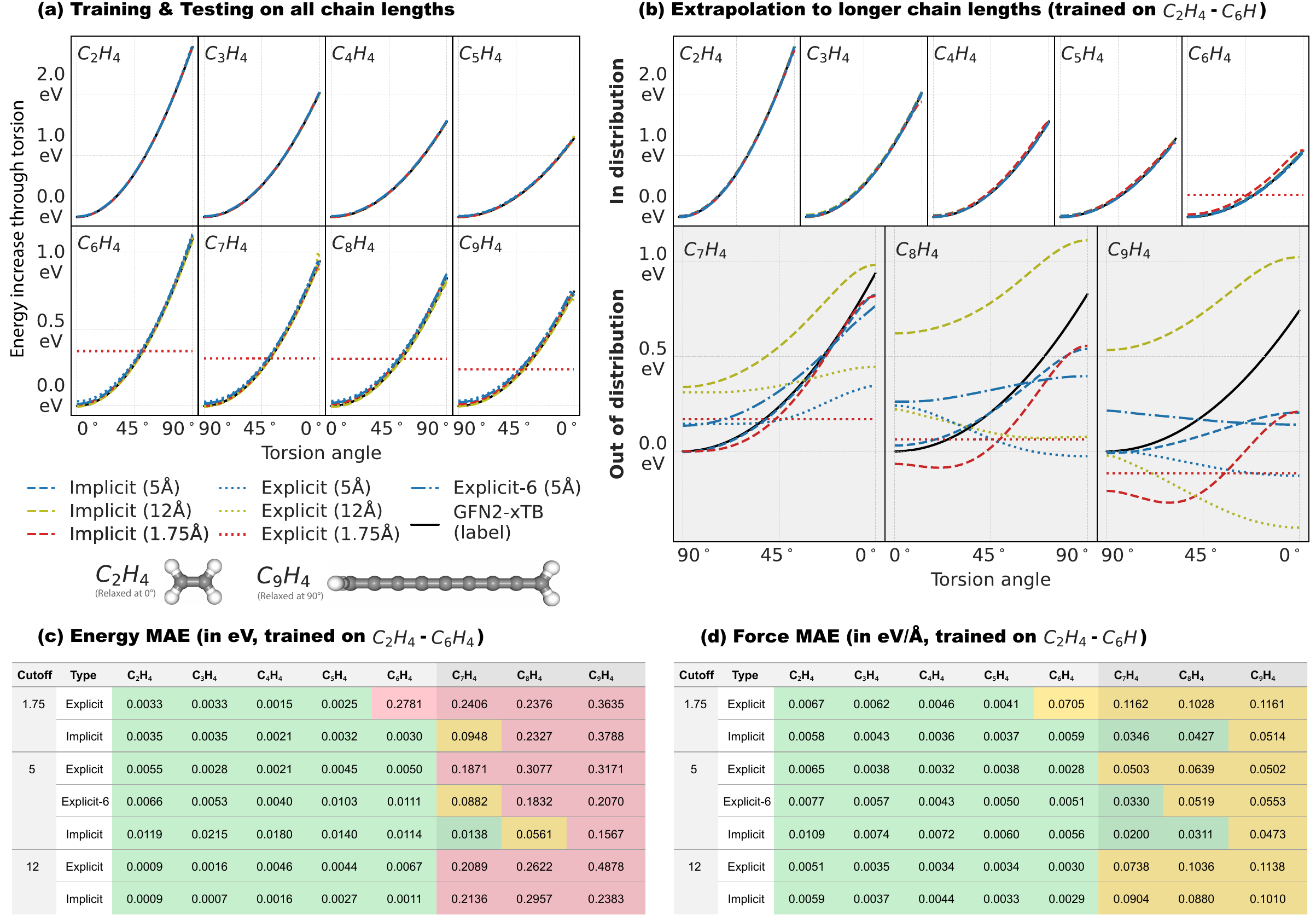}
    \caption{
    \textbf{Long-range generalization on the cumulene molecules \ce{C_nH4} ($n=2,\dots,9$).}
    Each panel shows the molecular energy relative to the relaxed structure as a function of the torsion angle between the two terminal \ce{CH2} groups; the black line is the GFN2-xTB label and colored lines are model predictions.
    Line colors denote the message-passing cutoff (\qtylist{1.75;5;12}{\angstrom}) and styles distinguish implicit (dashed) from three-layer explicit (dotted) models, with Explicit-6 (dash-dotted) a six-layer explicit baseline.
    \textbf{(a)} models trained on all eight molecules; almost all implicit and explicit variants reproduce the test curves, the explicit \qty{1.75}{\angstrom} model being the notable exception (constant prediction on the longer chains).
    \textbf{(b)} extrapolation, with models trained only on \ce{C2H4} to \ce{C6H4} and evaluated on the larger, unseen chains \ce{C7H4} to \ce{C9H4} (bottom row).
    Implicit models generalize comparatively well to the unseen molecules, whereas explicit models fit them poorly and predict a qualitatively wrong energy decrease on the longest chains.
    \textbf{(c \& d)} per-molecule and per-model mean absolute errors of energy (in \si{\electronvolt}) \& force predictions (in \si{\electronvolt\per\angstrom}) in the extrapolation setting, the columns \ce{C7H4} to \ce{C9H4} are unseen during training.
    Green cells achieved chemical accuracy
    of \qty{1}{\kcal\per\mole} = \qty{0.043}{\electronvolt} on the energy,
    or an analogous target force error
    of \qty{1}{\kcal\per\mole\per\angstrom} = \qty{0.043}{\electronvolt\per\angstrom}~\cite{chmiela2023}.
    Yellow cells are above $1\times$ and red cells above $3\times$ these target accuracies.
    Implicit modeling improves both energy and force profiles.
    }
    \label{fig:cumulene_all}
\end{figure*}

An implicit model is equivalent to an explicit MLFF with an infinite stack of weight-sharing layers, and therefore has no finite horizon on how far apart two structures can interact in its prediction.
In contrast, an explicit network can only propagate information over its `effective cutoff', at most the number of layers times the neighborhood cutoff ($3 \times \qty{5}{\angstrom}$).
We test whether this `infinite' effective cutoff, by communicating information across the whole system, improves predictions that depend on long-range effects, using the cumulene dataset~\cite{unke2021machine}.

The eight molecules \ce{C_nH4} ($n=2,\dots,9$) differ structurally only in the length $n$ of the double-bonded carbon backbone (Fig.~\ref{fig:cumulene_all}).
Within each molecule's test set only the torsion angle between the two terminal methylidene (\ce{CH2}) groups changes gradually, driving an increase in potential energy through a long-range electronic interaction.
The longest interatomic distance -- and thus the length scale over which this torsion must be communicated to predict the energy -- grows with the chain, from \qty{3.1}{\angstrom} on \ce{C2H4} to \qty{11.5}{\angstrom} on \ce{C9H4}.
Training on 4500 samples of each molecule allows almost all implicit and explicit models to predict the test curve well (Fig.~\ref{fig:cumulene_all}a).
The exception is a three-layer explicit network with a \qty{1.75}{\angstrom} cutoff. This cutoff is chosen to lie between the largest bonded (\qty{1.6}{\angstrom}) and smallest non-bonded (\qty{1.8}{\angstrom}) atom distance so that messages travel only along bonds.
This short-range model fails once aggregating the methylene positions takes more than three steps: beyond five carbons its prediction is constant, near the per-molecule training mean.
In contrast, an implicit model with the same \qty{1.75}{\angstrom} cutoff predicts the energy near-perfectly at any backbone size, without oversmoothing~\cite{li2018deeper,oono2020graph} even when up to ten propagation steps are required.

We use the same dataset to demonstrate extrapolation capabilities of implicit models.
By training only on cumulene molecules with up to six carbon atoms, we demonstrate that simply `turning MLFFs implicit' without changing the underlying architecture improves their generalization to the larger chains of up to nine carbons.
Dashed lines in the bottom row of Fig.~\ref{fig:cumulene_all}b highlight that all implicit models fit the test set energy curves of unseen molecules comparatively well.
Especially the default configuration with a message-passing cutoff of \qty{5}{\angstrom} performs well with energy errors around \qtyrange{0.04}{0.05}{\electronvolt} across the three unseen chains.
Every implicit model performs better than its explicit counterpart (Fig.~\ref{fig:cumulene_all}c).
Implicit energy errors occur most at the difficult high torsion region where the ground-truth energy has a discontinuous cusp and a physical bond break might occur. In part energy errors can be corrected by a constant energy offset which might be admissible as no systems of such large size were seen during training.
This shows up in the forces -- the energy's derivative, and the quantity that actually drives molecular dynamics -- which the implicit models recover more faithfully than their explicit counterparts (Fig.~\ref{fig:cumulene_all}d), even where the absolute energy is shifted.
Explicit models, in contrast, fit \ce{C7H4} poorly and give a qualitatively wrong prediction on \ce{C8H4} and \ce{C9H4}, modeling a decrease in potential energy while the ground-truth increases.
These findings hold across the ablated cutoffs (\qtylist{5;12;1.75}{\angstrom}); notably the larger \qty{12}{\angstrom} cutoff, which extends the explicit effective cutoff, hurts rather than helps generalization.
The only visible direction of improvement is seen when increasing the layer count of the explicit model from three to six layers (Explicit-6, blue dash-dotted line), which moves closer to the `infinite-depth' implicit modeling regime. The deep explicit model generalizes decently to chains with seven or eight carbons, but still predicts an incorrect energy decrease on the largest molecule.

For an explicit model of any fixed depth one can construct a system whose electronic effects act over too long a range to be modeled.
Implicit networks instead have an effective range that adapts to the geometry, and we observe no such breakdown across the chain lengths tested -- suggesting that the inductive bias of implicit modeling improves generalization beyond what a larger explicit cutoff or depth provides.

%% file: chapters/9b_Spectra.tex
\begin{figure}[htbp]
    \centering

    \begin{minipage}[b]{0.55\textwidth}
        \vspace{0pt} 
        \centering
        \includegraphics[width=0.9\linewidth]{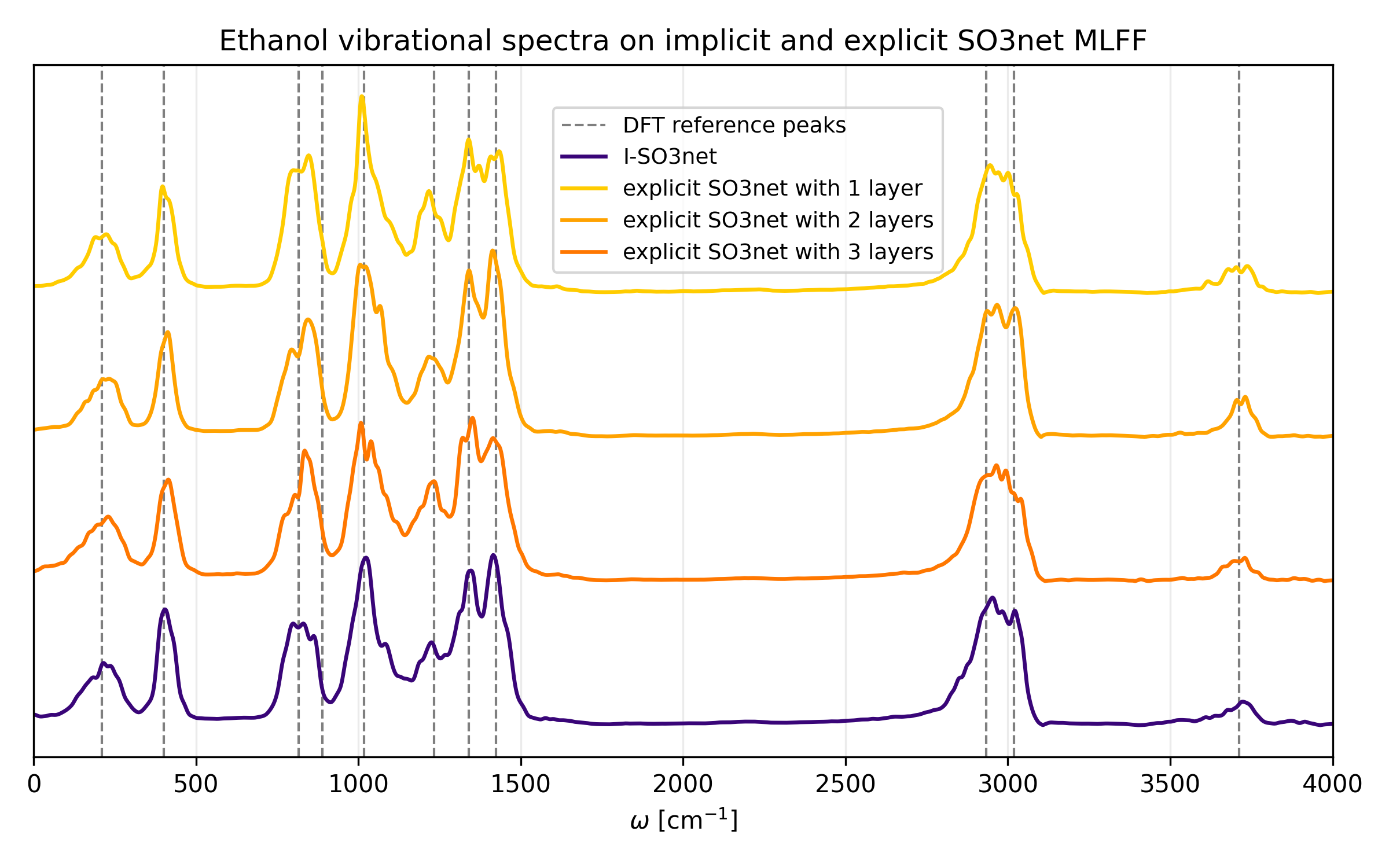}
        \caption{\textbf{Vibrational spectra for ethanol.} The I-SO3net model predicts the \qty{50}{\pico\second} trajectory with, on average, 1.77 iterations per timestep.
        Dashed lines indicate peaks of ethanol's vibrational spectrum in the MD17 reference trajectory.
        }
        \label{fig:Vibrational_spectrum_SO3net}
    \end{minipage}
    \hfill
    \begin{minipage}[b]{0.4\textwidth}
        \vspace{0pt} 
        \centering
        \begin{tabular}{r>{\hspace{0.8em}}r>{\hspace{0.8em}}r}
            
            \toprule
            Model & \makecell{Matched\\peaks} & \makecell{Mean deviation\\(\si[per-mode=power]{\per\centi\meter})} \\
            \midrule
            I-SO3net & 10 & 9$\pm$8 \\
            1 layer & 10 & 11$\pm$12 \\
            2 layer & 9 & 14$\pm$15 \\
            3 layer & 8 & 13$\pm$12 \\
            \bottomrule

        \end{tabular}
        \vspace{20pt}
        \captionof{table}{\textbf{Accuracy of spectral peaks.} This table qualitatively compares the performance of explicit SO3nets to an implicit SO3net on recreating physical observables.}
        \label{tab:spectra_peak_table}
    \end{minipage}

\end{figure}

\section{Vibrational spectra}
\label{sec:xiv_vibrational_spectra}

Efficient implicit force fields, with less than 2 layer evaluations on average, maintain the ability of explicit models to reproduce physical observables like vibrational spectra, as visualized in Fig. \ref{fig:Vibrational_spectrum_SO3net} and surpass their accuracy in certain spectral peaks.
Table \ref{tab:spectra_peak_table} helps to analyze the spectra more quantitatively. 
Peaks are detected as all frequencies with a higher intensity than their surrounding \SIwn{40} sliding window.
For a peak to `match' its DFT reference, it must be within this window.
The I-SO3net model captures the reference peaks most precisely, resulting in the highest number of matched peaks. Furthermore, it shows the lowest mean absolute deviation and variance from the DFT reference peaks.
While the explicit SO3net with one layer matches the same number of peaks as I-SO3net, it also exhibits many spurious oscillations, especially around the double peaks at \SIwn{1400} and \SIwn{3000} and the O-H peak at \SIwn{3700}.

%% file: chapters/9c_streamplot_details.tex
\begin{figure}[H]
\centering
\includegraphics[width=0.99\textwidth]{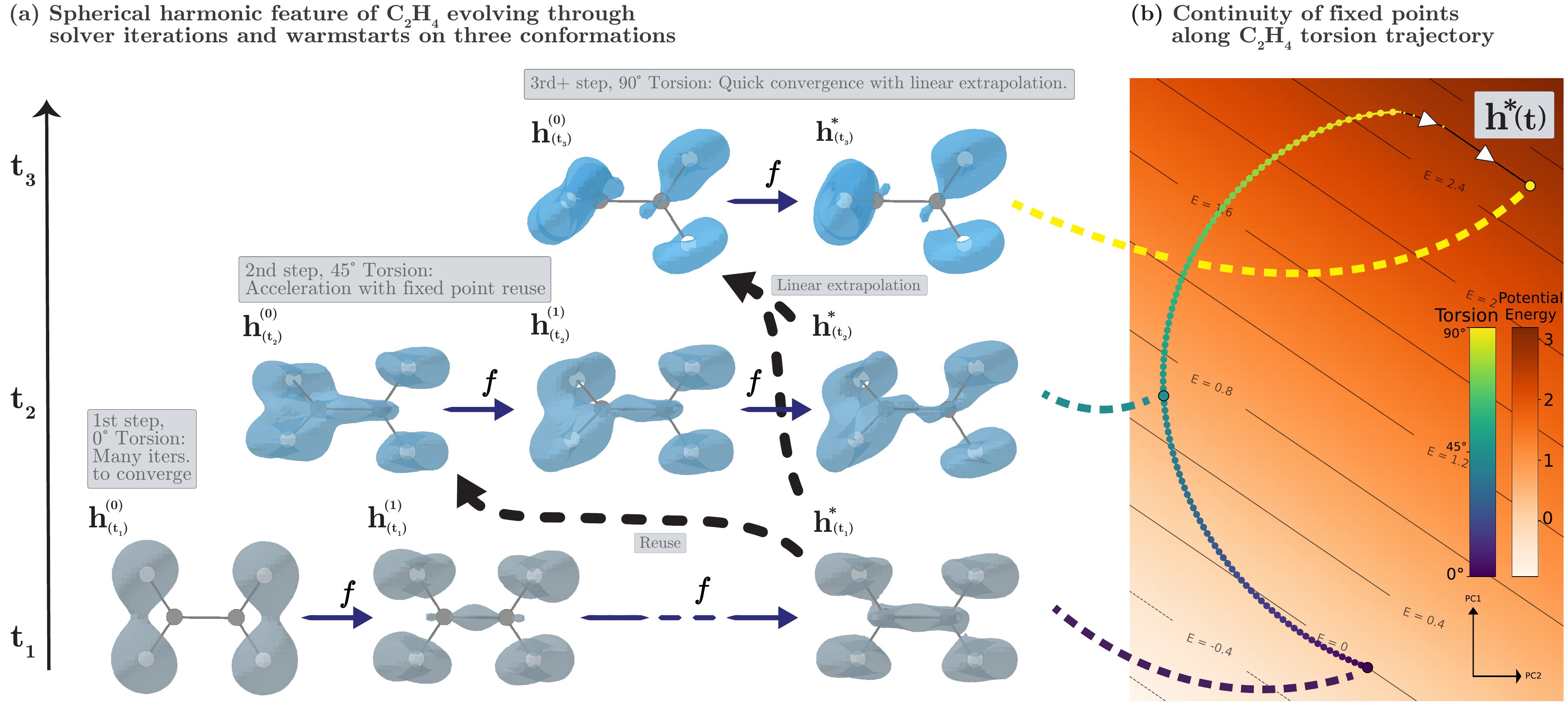}
\caption{
    \textbf{
Supplement to Fig.~\ref{fig:intro}(a)
on the \ce{C2H4} molecule.
}
(a)
Visualization of one spherical harmonics feature as a 3D density contour.
The bottom row shows its geometry-invariant initialization from the atomic numbers and how successive iterations of $f$ refine it to the fixed point.
The previous fixed point is applied to the new geometry in the second row (following the black dashed arrow),
warmstarting and accelerating the solver.
Linear extrapolation from two preceding fixed points (the converging black dashed arrows, top left) warmstarts the third step at $90^\circ$ torsion precisely.
(b)
Continuity of fixed-point trajectory:
The fixed-point trajectory of \ce{C2H4} projected onto the plane spanned by its first two principal components, overlaid on a contour plot of the potential energy surface sampled on a fine grid.
}
\label{fig:si-cumulene-supplement}
\end{figure}

\begin{figure}[H]
\centering
\includegraphics[width=0.8\textwidth]{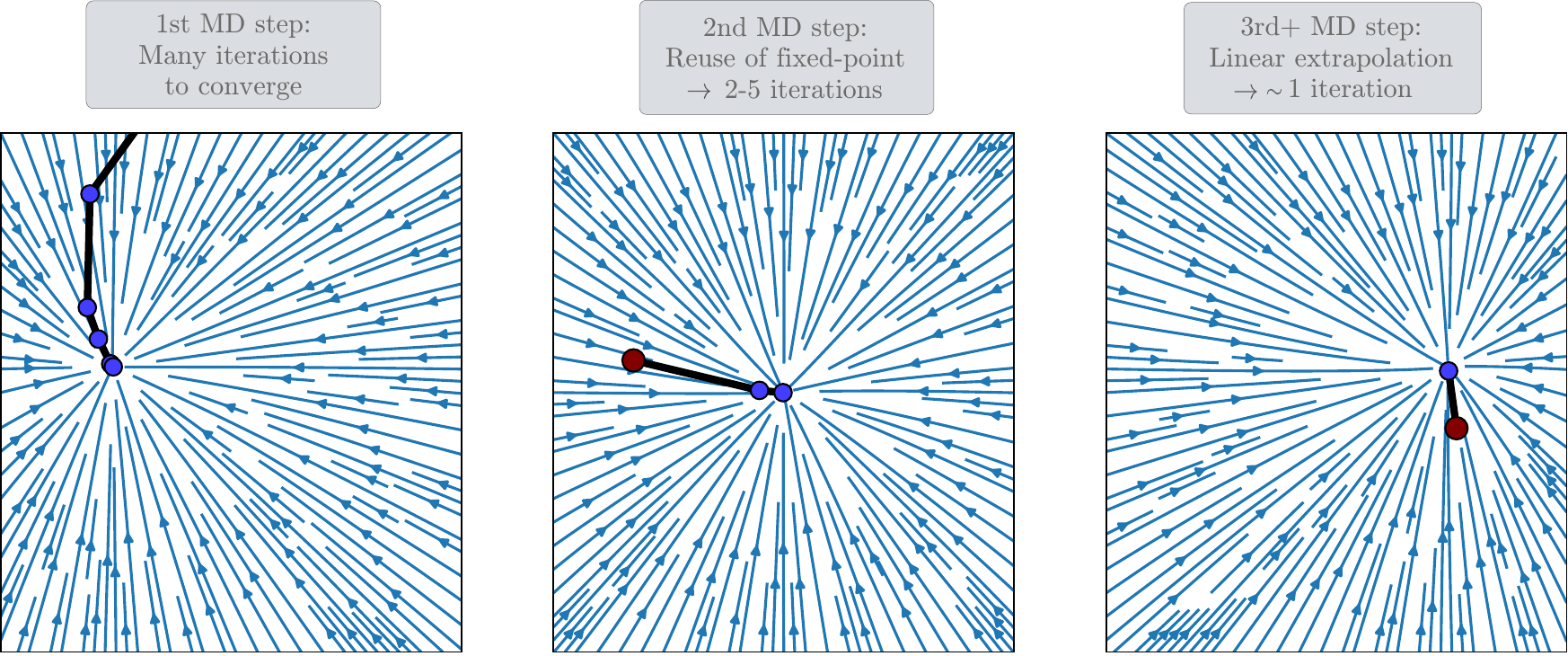}
\caption{
\textbf{
Vector fields underlying the 3D streamplot \mbox{Fig.~\ref{fig:intro}(b)}.
}
Three consecutive MD timesteps of I-PaiNN on azobenzene, showing the vector field (evaluated on a fine grid on the panel plane), warmstart initializations (red), and fixed-point iteration trajectories $f$ (blue). 
The plane visualized in each panel is spanned by the three fixed points of the three steps.
}
\label{fig:si-streamplot-supplement}
\end{figure}

\newpage
\section{Supplement to \mbox{Fig.~\ref{fig:intro}a,b} -- Fixed-point trajectory details}
\label{sec:si_fig1}

In the main text \mbox{Fig.~\ref{fig:intro}(a,b)} sketches the continuity and our extrapolation on the trajectory of fixed points $\hs(t)$ through time.
To visualize continuity we choose a torsion trajectory of \ce{C2H4} because the simplicity of the dynamics and the small size of the molecule allow for easy visualization.
In the main text we stylize the fixed-point trajectory for reasons of space and style (preserving the core characteristics).
Fig.~\ref{fig:si-cumulene-supplement}
supplements the visualization of the exact \ce{C2H4} fixed-point trajectory projected on the plane spanned by the two principal components.
We also sample the energy readout of the plane in a fine grid revealing a contour plot of the potential energy surface (PES).
The visualization is slightly flawed: it looks like torsion causes an initial dip in the energy prediction -- it does not.
The high-dimensional trajectory has monotonically increasing energy, only the two-dimensional slice on which the contour plot is visualized has an energy decrease in the place where the trajectory's projection lands.
Still, we can discern inner workings of the model:
At around $45^{\circ}$ the trajectory aligns with the first principal component, and with the energy gradient,
which causes the steep increase in predicted energy starting from there on (see SI Fig.~\ref{fig:cumulene_all}).

The torsion changes on the \ce{C2H4} are very small, making warmstarts and consecutive convergence too easy for an educational visualization.
All vector fields, warmstarts, and iteration $f$ shown in \mbox{Fig.~\ref{fig:intro}b} therefore visualize real data from an I-PaiNN model on the azobenzene system.
\mbox{Fig.~\ref{fig:si-streamplot-supplement}}
provides a more detailed view of the three panels (which are stacked vertically in the main text \mbox{Fig.~\ref{fig:intro}(b)}), including the vector field (blue, calculated on a fine grid on the plane) and the iterations $f$ (red; in the second and third timestep they start in the middle of the panel because of the warmstarts).
The plane visualized in all three panels (here and in the main text) is the plane spanned by the three fixed points of the three steps.

%% file: chapters/9d_energy_to_iteration.tex
\section{Supplement to Fig.~\ref{fig:results-iteration-depth} -- Increased solver iterations at high potential energies}
\label{sec:si_adaptive_iteration_depth}
\begin{figure}[htbp]
\centering
\includegraphics[width=\textwidth]{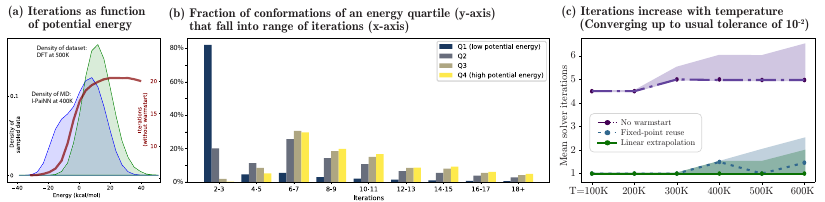}
\caption{\textbf{Dependence of average solver iterations on potential energy and temperature on Ac-Ala3-NHMe} (cf. Fig.~\ref{fig:results-iteration-depth}).
(a) mean iteration count (red, right axis) as a function of potential energy, overlaid on the energy density of the test MD at \qty{400}{\kelvin} (blue) and the DFT training dataset at \qty{500}{\kelvin} (green).
(b) fraction of conformations in each energy quartile that fall into each iteration bucket.
(c) Converging to a threshold of $10^{-2}$ takes ${\sim}5$ iterations without warmstarts and ${\sim}1$ iterations with either type of warmstart, with only slight increases with the temperature. To differentiate the warmstarts better, Fig.~\ref{fig:results-iteration-depth}d in the main text plots iterations to converge to a very fine threshold of $10^{-3}$.
}
\label{fig:si-ala3-supplement}
\end{figure}

In the main text, Fig.~\ref{fig:results-iteration-depth} shows how the iteration count of implicit models (especially without warmstarts) increases visibly in regions of the Ramachandran plot that are known to be outside of the main energy minima of the potential energy surface.
Fig.~\ref{fig:si-ala3-supplement} supplements this analysis by focusing directly on the relationship of the potential energy of the system (or its instantaneous temperature) and the iterations that the implicit model invests to solve it.
In Fig.~\ref{fig:si-ala3-supplement}a
the distribution of potential energies sampled in the MD of the implicit models is shown in blue.
We see that the MD22 trajectory from which the training set of the model is drawn samples higher energies (green):
As MD22 is simulated at \qty{500}{\kelvin} and our MD at \qty{400}{\kelvin}, MD22 had more kinetic energy in the system, which (by equipartition) caused statistically higher potential energies, too.
Notably, iterations, averaged over the 30 equiwidth bins of the plot, increase monotonically with potential energy except for noise caused by low sample counts at the edges of the blue distribution.
In Fig.~\ref{fig:si-ala3-supplement}b we take a more detailed view.
We categorize conformations into four quartiles based on their potential energy and color them accordingly.
Next, we sort conformations by how many iterations the solvers took to predict their interatomic forces.
Over \qty{80}{\percent} of the conformations in the lowest energy quartile are solved within fewer than 3 iterations.
For comparison, the overall average iteration count is 8.9, three times as high.
In contrast, high potential energy conformations are strongly skewed to take higher iteration counts, suggesting that the model invests more compute in conformations of higher chemical diversity.

Fig.~\ref{fig:si-ala3-supplement}c plots the iteration counts needed to reach our usual threshold of $10^{-2}$ in MD simulations run at different temperatures.
High temperatures indicate high kinetic energy of the molecule, which is statistically linked (through equipartition) to increased potential energies.
The correlation of iterations to potential energy is thus as expected:
increases in temperature correlate with a minor increase in iteration counts.

However, as Ac-Ala3-NHMe is very easy to solve using implicit models, the iteration counts of the warmstarted solvers stay very low, close to the minimum of one iteration per MD step at most temperatures.
Fig.~\ref{fig:results-iteration-depth}d in the main text plots convergence to a finer tolerance of $10^{-3}$ where linear extrapolation distinguishes itself, maintaining low iteration counts even at high temperatures.

%% file: chapters/9e_accuracy_tables.tex
\section{Accuracy tables}
\label{sec:si_accuracy_tables}

Table~\ref{tab:accuracy_by_molecule} displays the test set accuracies of implicit and explicit MLFFs for each molecule and model type, separated into three subtables for the PaiNN, SO3net, and SchNet architectures.
Only the force MAE for the implicit model using the equivariant merged layer norm (\textsc{LN})~\cite{liao2026equiformerv3} is shown in \si{\kcal\per\mole\per\angstrom}.
All other models on this architecture and molecule reference its performance and plot the percentual in- or decrease of their force MAE in shades from red to green.
Implicit models using the simpler equivariant \textsc{UL} normalization slightly underperform \textsc{LN} on close to all tasks, likely because of the lower expressiveness of the \textsc{UL} norm.
We train explicit models up to three layers on SO3net and SchNet, and up to six and eight layers with PaiNN on molecules of the MD22 and MD17 datasets, respectively.
Explicit models that share parameters across all layers are plotted for PaiNN on MD17 in a \maybe{\emph{gray, italic}} font and seem to mirror their explicit counterparts' performance closely.
Notably, the implicit model exceeds or matches the accuracy of its explicit counterparts up to five or six layers, while requiring only a fraction of their memory and compute using warmstarts.
The only exception forms benzene, a very stable molecule with little variation in nuclear positions and a high number of symmetries in its circular structure: It is solved best by a two-layer explicit PaiNN but degrades again on high depths, possibly because of the oversmoothing effect~\citep{li2018deeper,oono2020graph}.
Explicit models seem to scale worse to high depths on larger molecules. 
We speculate that this is at least in part caused by training dynamics getting increasingly unstable with depth, an issue that we could not fix reliably even when restarting training with different random seeds. Implicit models with their conceptually infinite depth avoid this issue, possibly by training and regularizing the model for a more restrictive fixed-point representation, or by adding input-injection and normalization to the architectures.

Table~\ref{tab:accuracy} summarizes the detailed tables by taking the mean over force MAEs of all molecules in a dataset and computing the relative in- or decrease over these aggregated scores.
Across the three architectures, explicit models on the larger molecules of MD22 often catch up to the implicit accuracy with about three layers, bringing the necessary compute down to roughly the level of an implicit warmstarted force field. 
The memory consumption of implicit models in this setting is precisely one third of a three-layer explicit network of the same architecture, driving significant savings especially on large molecular systems, cf. Fig.~\ref{fig:si-memory-lineplot}.

\begin{table}[p]
    \centering
    \includegraphics[scale=0.71]{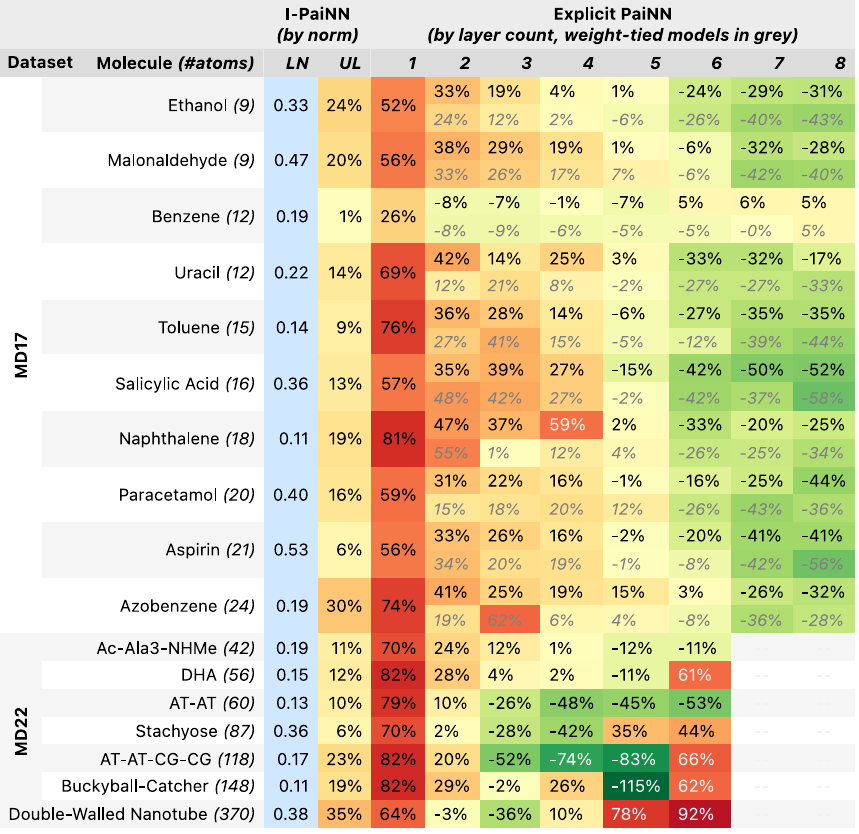}
    \includegraphics[scale=0.71]{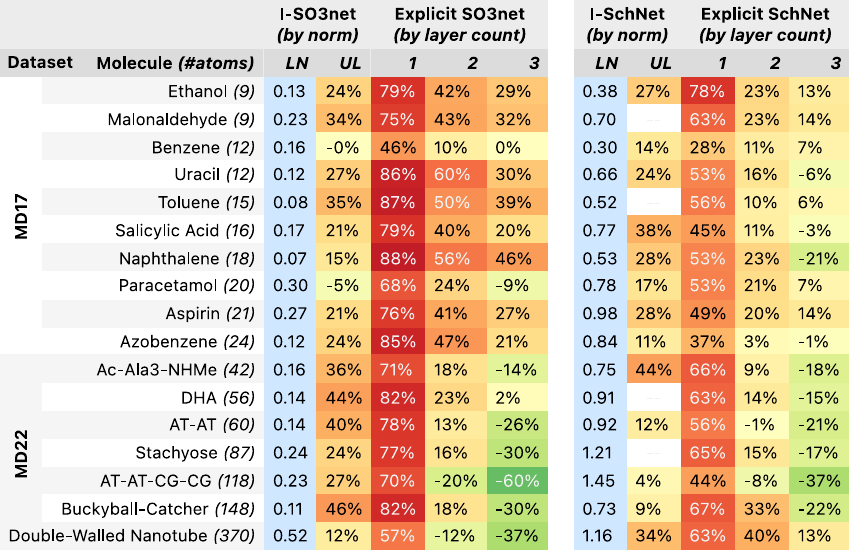}
    \caption{
        Force accuracy per molecule, with three sub-tables focusing on the PaiNN, SO3net, and SchNet architectures.
        In all tables the force MAE of the implicit model using the equivariant layer norm (\textit{LN}) is shown in \si{\kcal\per\mole\per\angstrom} on a blue background.
        All other models show their increase (red) or decrease (green shades) in force MAE relative to the \textit{LN} baseline.
    }
    \label{tab:accuracy_by_molecule}
    
    \bigskip
    
    \centering
    \includegraphics[scale=0.71]{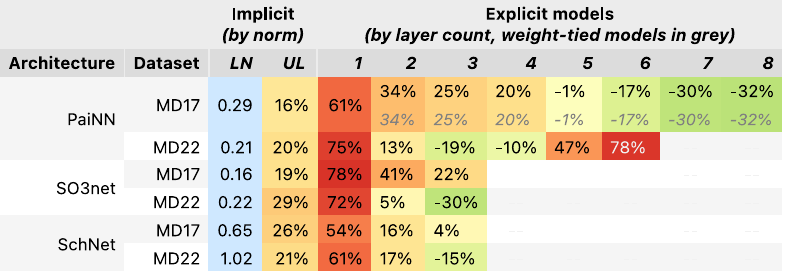}
    \caption{
        Force accuracy per architecture and dataset, averaging the per-molecule force MAE shown above and using the same color scheme.
        The implicit \textsc{LN} is shown in \si{\kcal\per\mole\per\angstrom};
        other models note their in- or decrease relative to \textsc{LN}.
    }
    \label{tab:accuracy}
\end{table}

\FloatBarrier

%% file: chapters/9f_memory_plot.tex
\section{Memory consumption}
\label{sec:si-memory}

\begin{figure*}[!t]
\centering
  \includegraphics*[width=\textwidth]{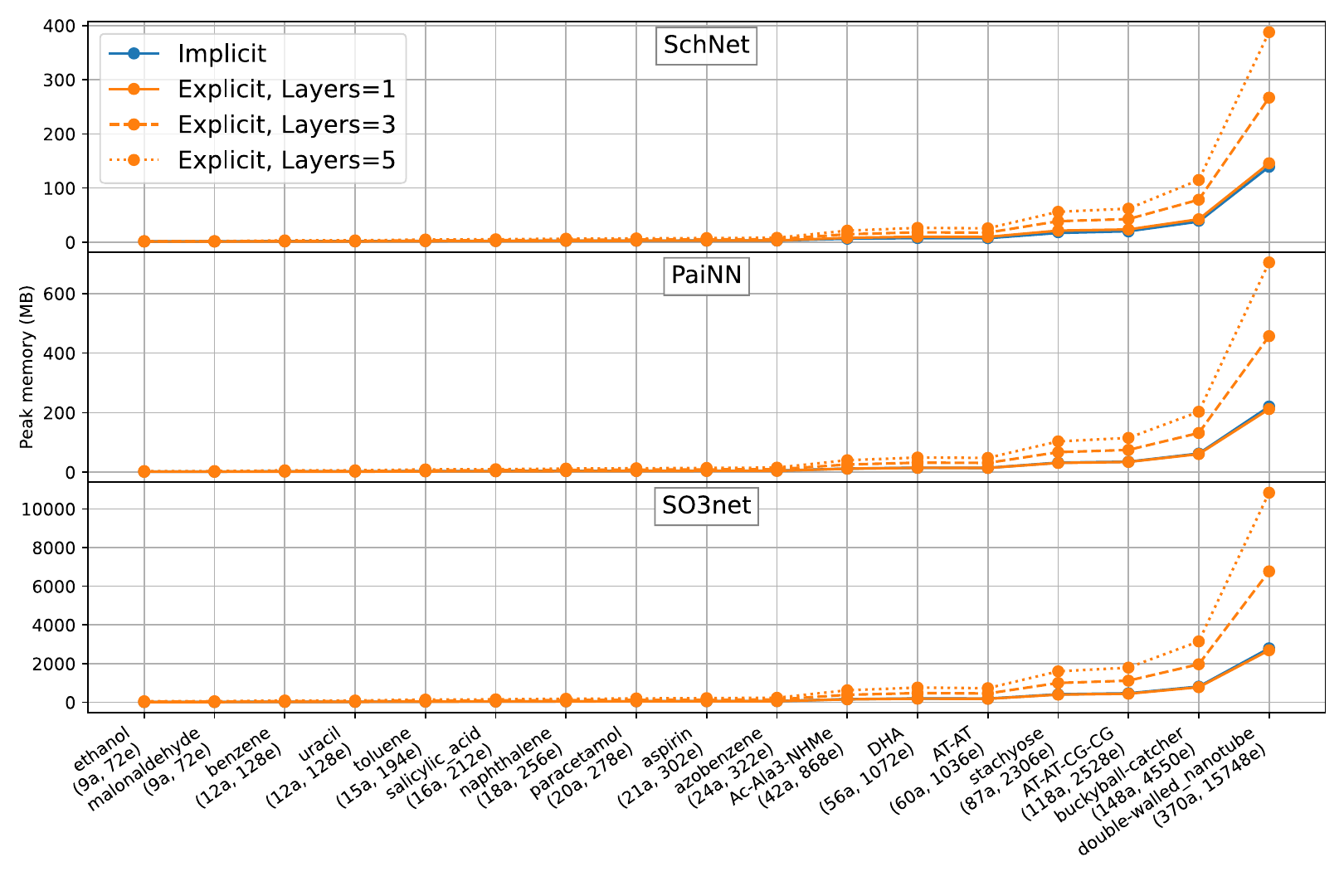}
  \caption{
    Peak memory required for calculating forces as an energy derivative in implicit and explicit models of one, three, and five layers.
    Implicit models require roughly 1/K of the memory of an explicit model with K layers, as only the last application of $f$ of the message-passing layer near the fixed point $\hs = f(\hs, \x)$ is required for implicit differentiation.
    The x-axis denotes the tested systems and their count of atoms (a) and message-passing edges (e) in the \qty{5}{\angstrom} cutoff, on a conformation that's randomly drawn from the datasets.
  }
\label{fig:si-memory-lineplot}
\end{figure*}



Fig.~\ref{fig:si-memory-lineplot} documents the memory consumption of an implicit model and three explicit models with 1, 3, and 5 message-passing layers on the SchNet, PaiNN, and SO3net architectures.
The x-axis denotes a range of molecules for which the measurement was performed, as well as the number of atoms and edges within a \qty{5}{\angstrom} cutoff that drive the memory consumption.

We use implicit differentiation to calculate forces as the derivative of the energy measured at the fixed-point representation \hs, which requires only local derivative information around the fixed point and not the entire fixed-point solver's path, cf. Sec.~\ref{subsec:methods_implicit_forces}.
In practice, this means that only the intermediate activations within the last iteration of the solver need to be kept in memory, no matter how many iterations are necessary to find the fixed point.
This contrasts with explicit force fields which need to store and backpropagate through all of their message-passing layers (irrespective of whether the layers share learned parameters).

In the largest measured system, the double-walled nanotube, this brings the memory consumption of a three- or five-layer explicit SO3net during force inference to \qty{6.8}{\giga\byte} or \qty{10.8}{\giga\byte}, respectively, while the implicit model requires only \qty{2.8}{\giga\byte}.
These measurements include the fixed-point history required for the linear warmstarts of implicit models. Their contribution and other factors such as the reduced parameter storage are minimal compared to the storage of message-passing activations.

%% file: chapters/9g_algorithm.tex
\section{Algorithm of Implicit Conservative Force Fields}
\label{sec:algorithm}

\begin{figure*}[h]
\centering
\small
\noindent
\begin{minipage}[t]{0.44\linewidth}
\hrule\vspace{0.6em}

\noindent\textbf{Algorithm 1:} \textsc{ImplicitMLFF} \\\textbf{Implicit energy and force evaluation}
\vspace{0.3em}

\input{algorithms/implicit-solve}

\vspace{0.6em}\hrule
\end{minipage}%
\hfill
\begin{minipage}[t]{0.54\linewidth}
\hrule\vspace{0.6em}

\noindent\textbf{Algorithm 2: Warmstarts \\ (Using a two-step history)}
\vspace{0.3em}

\input{algorithms/warmstart-implicit_solve}

\vspace{0.6em}\hrule

\vspace{0.8em}
\small
\sloppy
\leftskip=0pt \rightskip=0pt \parfillskip=0pt plus 1fil
\noindent
\textbf{Algorithm 1}
stores only the computation graph of the last iteration
$(\h_\text{prev}, \R) \xmapsto{f} \h$.
The backward solver evaluates the 
VJP
of this map
wrt. both its inputs
to solve equations~\ref{eq:results_v_star} and \ref{eq:results_force_readout_derivative} for \F and \vbars in parallel.
\par\vspace{0.4em}\noindent
\textbf{Algorithm 2} stores a rolling history of the previous two timesteps fixed points for \hs and \vbars and warmstarts Algorithm 1 on the next timestep with their linear extrapolation.
\par
\end{minipage}
\end{figure*}

Algorithm 1 summarizes the entire implicit energy-conservative MLFF.
After receiving warmstarts for \h and \vbar from Algorithm 2, it iteratively solves for both \hs to obtain the system's potential energy $E$,
and then for \vs to derive energy-conserving forces $\F = \nabla_{\R} E$.

Notably, 
it decomposes the self-consistency equation~\eqref{eq:results_v_star} of the fixed-point adjoint $\vs = \vbars + \w$ into two parts,
an intermediate variable 
$\vbars := (\frac{\partial f}{\partial \hs})^\top \vs$
and the energy-head adjoint
$\w := \frac{\partial f_E}{\partial \hs}$.
We capture the intermediate
\vbars
on preceding MD steps
and extrapolate from them
analogously to equation~\eqref{eq:results_h0}
for a guess of the current MD step
$\vbar \approx \vbars$.
This allows Line 10 to estimate the fixed-point adjoint
$\v \gets \vbar + \w$
in a single vector addition
and to compute the improved estimates for
$\vbar \gets \vs \frac{\partial f}{\partial \hs}$
and
$\F \gets \vs \frac{\partial f}{\partial \R}$
in parallel,
with a Vector Jacobian Product (VJP)
of the stored last execution of $f$.
Within a few iterations
we obtain accurate forces \F, as well as a nearly converged $\vbar \approxeq \vbars$ 
from which the next MD step is warmstarted.

Compared to serially deriving \vs in equation~\eqref{eq:results_v_star}
and then \F in equation~\eqref{eq:results_force_readout_derivative},
parallel derivation using \vbar saves one VJP per MD step.
Since implicit force fields often require only one forward and backward iteration each to converge sufficiently after warmstarts,
and $f$ and its VJP have a similar time complexity~\cite{griewank2014automatic},
saving one VJP can reduce the per-step cost of the entire implicit MLFF by up to \qty{33}{\percent}.

Finally, if $f$ involves any computation regarding solely \Z or \R and not the iterate \h, they should be computed once into intermediate embeddings, say $\h_{\Z}$ and $\h_{\R}$, and fed into the remaining $f$ at every iteration without recomputing them.
Similarly during force derivation, each backward iteration should yield the cotangent $dE/d\h_{\R}$, but only once convergence is detected should we backpropagate through the embedding from $\h_{\R}$ to \R to derive atomic forces. This is analogous to cross-layer optimization in explicit networks.

%% file: algorithms/implicit-solve.tex
\begin{algorithmic}[1]
\Require Atomic positions \R, atomic numbers \Z,
\Statex \hspace{\algorithmicindent}
        initial state \h and adjoint \vbar,
        tolerances $\varepsilon_{\h},\varepsilon_{\vbar}$
\Ensure Energy $E$, forces \F, and converged \h, \vbar

\Statex \hfill
\makebox[0pt][r]{%
\begin{tabular}{r}
\Comment{\text{Forward}}\\
\text{(Calculate fixpoint \h and $E$)}
\end{tabular}}

\Repeat
  \State $\h_{\mathrm{prev}} \gets \h$
  \State $\h \gets f(\h_{\mathrm{prev}}, \R, \Z)$
\Until{$\lVert \h - \h_{\mathrm{prev}} \rVert \le \varepsilon_{\h}$}
\State $E \gets f_E(\h)$

\Statex \hfill
\makebox[0pt][r]{%
\begin{tabular}{r}
\Comment{\text{Backward}}\\
\text{(derive \F implicitly at fixed point $\h$)}
\end{tabular}}

\State $\mathbf{w} \gets \nabla_{\h} E$
\Repeat
  \State $\vbar_{\mathrm{prev}} \gets \vbar$
    \State $\v \gets \vbar + \mathbf{w}$
    \State $( \vbar, -\F ) \gets 
    \v^\top 
    {\partial f} / {\partial (\h_{\mathrm{prev}}, \R)}$
\Until{$\lVert \vbar - \vbar_{\mathrm{prev}} \rVert \le \varepsilon_{\vbar}$}

\State \Return $(E,\F, \h, \vbar)$
\end{algorithmic}

%% file: algorithms/warmstart-implicit_solve.tex
\begin{algorithmic}[1]
\Stateful Previous fixed points $(\hs_{-\Delta t},\hs_{-2\Delta t}), (\vbars_{-\Delta t},\vbars_{-2\Delta t})$
\Require Atomic positions \R, atomic numbers \Z,
\Ensure Forces \F
        
\Statex \Comment{Warm-start via linear extrapolation}
\State $\h \gets 2\,\hs_{-\Delta t} - \hs_{-2\Delta t}$
\State $\vbar \gets 2\,\vbars_{-\Delta t} - \vbars_{-2\Delta t}$

\Statex \Comment{Implicit fixed point + differentiation}
\State $(E,\F,\h,\vbar) \gets \textsc{ImplicitMLFF}(\R,\Z,\h,\vbar)$

\Statex \Comment{Commit for next call }
\State $(\hs_{-2\Delta t},\hs_{-\Delta t}) \gets (\hs_{-\Delta t},\h)$
\State $(\vbars_{-2\Delta t},\vbars_{-\Delta t}) \gets (\vbars_{-\Delta t},\vbar)$

\State \Return \F
\end{algorithmic}

%% file: chapters/9h_warmstarts_ab.tex
\section{Extrapolation from three or more fixed points}
\label{sec:si-warmstarts-extended}

The linear predictor in equation~\eqref{eq:results_h0} can be extended to higher polynomial degree at negligible additional cost.
A Taylor expansion of the equilibrium trajectory $\hs(t)$ around time $t$ gives the degree-$k$ approximation
\begin{equation}
T_k(t + \Delta t) = \sum_{j=0}^{k} \frac{1}{j!} \frac{d^j \hs}{dt^j}\bigg|_{t} \Delta t^{\,j}.
\label{eq:si_taylor}
\end{equation}
The key idea behind Adams--Bashforth (AB) methods~\cite{hairer1993solving} is that $k$ samples of a function's first derivative at consecutive time steps -- computed from $k{+}1$ stored values, i.e.\ $k$ predecessors -- implicitly encode all derivatives up to order $k{-}1$, via the finite differences between samples.
Rather than computing the higher derivatives in equation~\eqref{eq:si_taylor} explicitly, AB fits a degree-$(k{-}1)$ polynomial interpolant through $k$ first-order derivative samples and integrates it one step forward, yielding the prediction:
\begin{equation}
\hs_{(t+\Delta t)} = \hs_{(t)} + \Delta t \sum_{j=0}^{k-1} b_j \, \bm{\varphi}_{(t - j\Delta t)}.
\label{eq:si_ab}
\end{equation}
Here, $\bm{\varphi}_{(t - j\Delta t)}$ denotes the first time derivative of \hs at time $t{-}j\Delta t$.
The coefficients $b_j$ are fixed rational numbers determined solely by the order $k$ (Table~\ref{tab:si_ab_coeffs}); no matrix inversions or least-squares fits are required.

\begin{table}[h]
\centering
\caption{Adams--Bashforth coefficients $b_j$ in equation~\eqref{eq:si_ab}.
For $j{\geq}k$, $b_j=0$.}
\label{tab:si_ab_coeffs}

\renewcommand{\arraystretch}{1.5}
\setlength{\tabcolsep}{3.5pt}
\setlength{\aboverulesep}{0pt}   
\setlength{\belowrulesep}{0pt}   

\begin{tabular}{@{}c l@{} r}
\toprule
$k$ & $b_j$ for $j{<}k$ \\
\midrule
1 & $1$ \\
2 & $\tfrac{3}{2},\; -\tfrac{1}{2}$ \\
3 & $\tfrac{23}{12},\; -\tfrac{16}{12},\; \tfrac{5}{12}$ \\
4 & $\tfrac{55}{24},\; -\tfrac{59}{24},\; \tfrac{37}{24},\; -\tfrac{9}{24}$ & \rule{0pt}{13pt} \\
\bottomrule
\end{tabular}

\end{table}

In the classical ODE setting, the derivatives $\bm{\varphi}_{(t - j\Delta t)}$ are available directly from the evaluation of the ODE's right-hand side.
In our setting, the solver returns fixed-point \emph{values} $\hs_{(t)}, \hs_{(t-\Delta t)}, \dots$ rather than time derivatives.
We estimate the required first-order time derivatives from finite differences of the stored values.
For the most recent point, only a backward difference can be computed, as $\hs\att[+]$ is not yet available:
\begin{equation}
\bm{\varphi}_{(t)} \approx \frac{\hs_{(t)} - \hs_{(t - \Delta t)}}{\Delta t}.
\end{equation}
For all earlier points $t - i\Delta t$ with $i \geq 1$, a central difference can be formed:
\begin{equation}
\bm{\varphi}_{(t - i\Delta t)} \approx \frac{\hs_{(t-(i-1)\Delta t)} - \hs_{(t-(i+1)\Delta t)}}{2\,\Delta t}.
\label{eq:si_central}
\end{equation}
Substituting these derivative estimates into equation~\eqref{eq:si_ab} gives the degree-$k$ warm-start for \hs, and analogously for \vbar.
At $k{=}1$, this reduces to the linear predictor of equation~\eqref{eq:results_h0}.
Higher degrees require storing $k{+}1$ previous fixed points but no additional model evaluations, making the computational overhead negligible.

%% file: chapters/9i_iteration_increases.tex
\section{Setup for iteration count measurements on aspirin}
\label{sec:experiment_iter_count_details}

To measure average iteration counts during MD on aspirin in Fig.~\ref{fig:results-1},
$10 {\times} \qty{5}{\pico\second}$ $NVE$ dynamics are simulated on the aspirin system at \qty{500}{\kelvin} for each tested model and warmstart configuration.
We set up all system geometries and momenta to be representative and have a broad coverage of the system ensemble.
For this, we first sample 10 different random conformations of the MD17's long aspirin trajectory and draw their atomic momenta from a normal distribution with a mean corresponding to the target temperature of \qty{500}{\kelvin}.
Finally, we apply three short equilibration phases,
running Langevin dynamics thermostats with coupling strengths
$t_{LTC} \in$ \SIlist{50;200;1000}{\femto\second}
for $t_{LTC}$ each.
This ensures that the total energy of the system corresponds to the target temperature, and momenta of every atom are pointing in a physically likely direction at the start of the measurement.
We find these starting positions and momenta
  to be representative of the model ensemble,
  as the gathered statistics are consistent across the 10 replicas
  and across the length of the MD simulations.
The fixed points \hs and \vbars are converged up to the required tolerance of $10^{-2}$ during the simulations,
  which leads to the visualization of residuals in 
  Fig.~\ref{fig:results-1} (C.\MakeUppercase{\romannumeral 2}.)
  to stop soon after crossing the red dotted threshold.

\section{Setup for stability measurements on Ac-Ala3-NHMe}
\label{sec:experiment_stability_details}

For the stability experiments in Fig.~\ref{fig:results-accuracy-stability}B--C that focus on non-conservative energy drift, we measure the average temperature increase (per \si{\femto\second}) across $15$ simulations over \qty{50}{\pico\second} on the Ac-Ala3-NHMe molecule.
We prepare each trajectory by relaxing a randomly drawn conformation from MD22, re-adding kinetic energy corresponding to a temperature of \qty{600}{\kelvin}, and bringing it into thermal equilibrium at about \qty{300}{\kelvin} using a short \qty{5}{\pico\second} NVE simulation at a fine solver tolerance of $10^{-4}$.

%% file: chapters/9j_ising.tex
\begin{figure}[h]
  \centering
    \includegraphics[width=.7\linewidth]
    {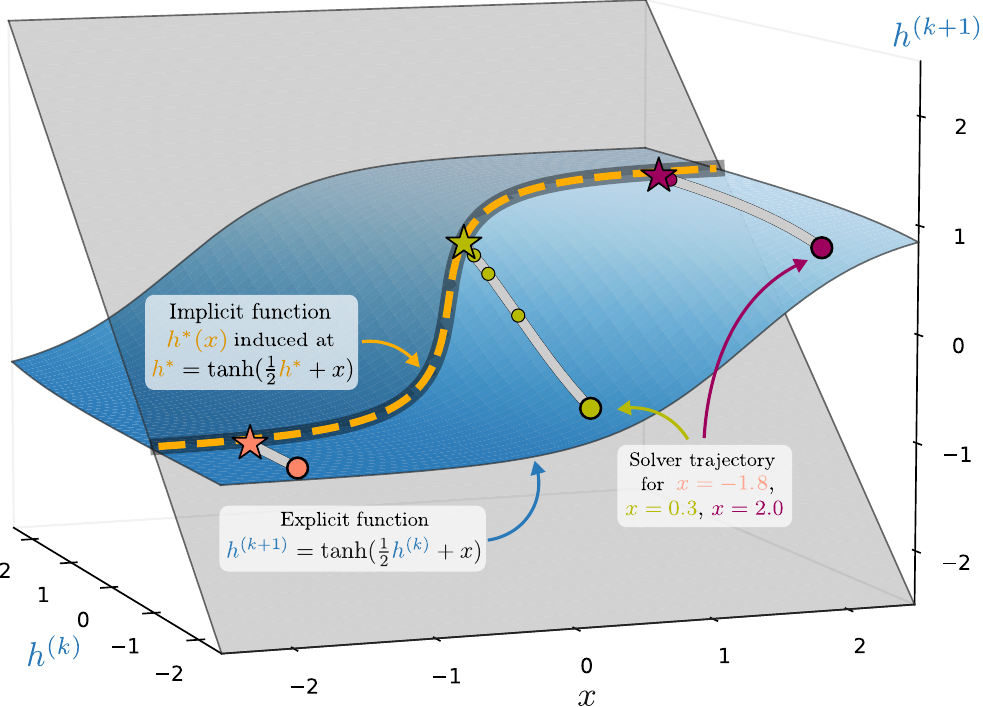}

\caption{\textbf{Full illustration of an implicit model with a scalar-valued fixed point:}
The magnetic Ising model (mean-field Curie-Weiss) with $h^*$ as the average magnetization, coupling strength $\frac 1 2$, and a varying external magnetic field $x$.
The explicit function 
$h^{(k+1)} {=} \tanh(\frac 1 2 h^{(k)} {+} x)$ -- analogous to $f$ in the main text -- 
is plotted in blue.
Where its input is equal to its output
$h^{(k+1)} {=} h^{(k)}$ (on the gray diagonal),
it induces the \emph{implicit map}
$x \mapsto h^*(x)$, which is visualized as the red dashed intersection.
No analytical expression for $h^*(x)$ exists,
yet the (implicit) function varies smoothly with its input parameter $x$.
The fixed point $h^*(x)$ carries physical meaning as it is the average magnetization of the spins
and has a well-defined derivative,
the magnetic susceptibility
$\chi = \frac{d}{d x} h^*(x)$.
This is analogous to the implicit energy prediction $E$ and implicit derivation of forces \F in equations~\eqref{eq:results_E_F} and~\eqref{eq:results_h_star}.
Because of its smooth shape, warmstarting with polynomial extrapolation like in equation~\eqref{eq:results_h0}
from neighbouring points of the red fixed-point trajectory $h^*(x)$
can accelerate convergence.
Three non-warmstarted solver trajectories (orange, olive, and wine) visualize how the convergence speed depends on the flatness of the underlying explicit function.
Jacobian regularization as introduced in Sec.~\ref{sec:model_training} aims to nudge learned parameters to regularize the spectrum of the blue plane, producing exactly such flat regions and fast convergence.
}
\label{fig:ising}
\end{figure}

\section{Implicit model illustration in 1D -- The magnetic Ising model}
\label{sec:si_ising}

Fig.~\ref{fig:ising} visualizes the full mechanics of a one-dimensional implicit model in one plot.
This includes most of the concepts discussed in the main text:
Solver iteration, 
the implicit function,
its smooth dependence on inputs, and a well-behaved spectrum as desiderata.

In general terms, an implicit function is any function whose input depends on its output.
As such, most physical laws can be described in terms of implicit functions --
often observations made in nature are produced as the balance, or equilibration, of opposing and interacting effects.
Examples include the equilibrium temperature of a planet, where absorbed solar flux and emitted thermal radiation form a self-consistent equation for the `fixed-point' temperature~\cite{budyko1969,mcguffie2005}.
In quantum chemistry, the Hartree-Fock and Kohn-Sham density functional theory equations are fixed-point problems where the molecular orbitals must diagonalize an operator that is itself constructed from those same orbitals~\cite{szabo1989,parr1989}.

We analyze the Ising model, a simple, well-studied \emph{implicit} model of the interaction of magnetic spins.
Specifically, in the mean-field Curie-Weiss formulation the whole system reduces to a one-dimensional (scalar) fixed-point equation. 
Adopting our notation, it models
the average magnetic dipole moment $h^*$ for all the many spins, under the condition that they are only influenced by the mean-field of all the other spins $h^*$ in the system and an external magnetic field $x$.
This self-consistency and input dependence mirror the structure of I-MLFFs precisely, and just as the fixed point of an I-MLFF carries physical meaning in the form of the total energy, the fixed point of the Ising model corresponds to the average magnetization per spin. Similarly, the implicit derivative of the fixed point equates to the magnetic susceptibility of the system. These and further analogies, such as warmstarting and the effect of the shape of the underlying explicit function, are described in the caption of Fig.~\ref{fig:ising}.

%% file: chapters/9k_HNN.tex
\section{Implicit Hamiltonian Neural Networks (I-HNN)}
\label{sec:xiii_HNN}

\begin{figure}[h]
    \centering
    \includegraphics[width=1\linewidth]{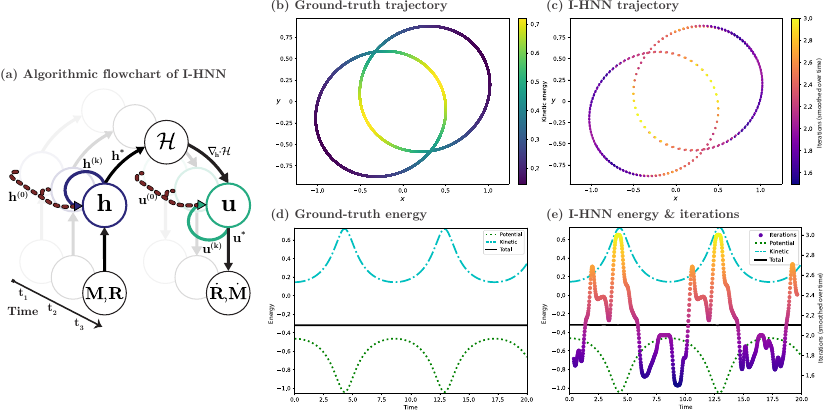}
    \caption{\textbf{Architecture and performance of an I-HNN on the gravitational two-body problem.} (a) a schematic visualization of the I-HNN similar to \ref{fig:intro}.  (b) the ground truth trajectory and the kinetic energy at each point. (c) the trajectory predicted by the I-HNN almost perfectly matches the ground truth and the iteration count correlates pointwise with the kinetic energy. (d) and (e) show the energy dynamics of the ground truth and the dynamics predicted by the I-HNN respectively and the iterations needed for the I-HNN to converge. For illustration purposes, the iteration count was smoothed symmetrically over 30 time steps with a Gaussian convolution.}
    \label{fig:HNN_trajectory}
\end{figure}

Implicit MLFFs have unique benefits in compute and memory costs as the sequential nature of predictions permits warmstarting fixed-point solvers, and the implicit derivation requires backpropagating only a single layer instead of an entire stack of neural network layers.
The same techniques generalize naturally to diverse applications in physical simulation and image processing, cf. Sec.~\ref{sec:discussion}, which we demonstrate here on a small Implicit Hamiltonian Neural Network (I-HNN) trained on the gravitational two-body task~\cite{NEURIPS2019_HNNs}.

In contrast to HNNs, MLFFs as discussed in this study predict the potential energy $E$ only from the particle positions \R (and some constant inputs),
and derive the forces $\F = -\nabla_{\R} E$. A separate integrator like Velocity-Verlet integrates the acting forces into the system momenta \M and drives the system positions.
This split is not possible for more general \emph{Hamiltonian} systems, like for a charged particle in a magnetic field where both the positions and momenta influence the potential energy $E(\R, \M)$ (as well as the kinetic energy $K(\R, \M)$).
Here, simulation does not require predicting the potential energy but the more general Hamiltonian
$\mathcal{H}=E+K$
and integrating the change in both of the system's inputs simultaneously, $[\dot{\R}, \dot{\M}] = [\nabla_{\M} \mathcal{H}, \nabla_{\R} \mathcal{H}] $.

Our modeling techniques extend naturally to this domain.
On the two-body gravity simulation, we adopt the small Multi-Layer Perceptron used as an HNN in~\cite{NEURIPS2019_HNNs} into an implicit structure analogous to I-MLFFs visualized in Fig.~\ref{fig:HNN_trajectory} (a):
First, a linear layer embeds both inputs into $\h_{\R, \M}$.
Then an affine layer ($f$) with a pointwise $\tanh$ non-linearity is iterated
$\h{k+1} = f(\h{k} + \h_{\R, \M})$
until the relative residual is smaller than $10^{-2}$.
The Hamiltonian $\mathcal{H}$ is predicted with a linear layer from \hs.
Implicit differentiation analogous to SI Sec.~\ref{sec:algorithm} gives the instantaneous changes 
$[\dot{\R}, \dot{\M}] = [\nabla_{\M} \mathcal{H}, \nabla_{\R} \mathcal{H}] $,
which are used to evolve the system in time.
Using linear extrapolation from two previous time steps to warmstart both the forward and backward solver, the I-HNN needs on average 2.1 layer calls to converge, while maintaining the total energy of the system.
Fig.~\ref{fig:HNN_trajectory} (c) and (e) illustrate how these iterations are distributed across the trajectory:
More compute is invested in the `difficult' regions of high kinetic energy (in which both implicit and explicit models tend to produce the most significant energy-conservation errors).